\pdfoutput=1

\documentclass[11pt]{article}

\usepackage[final]{acl}

\usepackage{times}
\usepackage{latexsym}

\usepackage[T1]{fontenc}

\usepackage[utf8]{inputenc}

\usepackage{microtype}

\usepackage{inconsolata}

\usepackage{graphicx}

\usepackage{amsmath} 
\usepackage{mathtools} 
\usepackage{amsfonts}
\usepackage{glossaries}
\usepackage{CJKutf8}
\usepackage{algorithm}
\usepackage{algpseudocode}
\usepackage{amsmath}
\usepackage{amssymb}
\usepackage{bm}
\usepackage{geometry}
\usepackage{hyperref}
\usepackage{enumitem}
\usepackage{multirow}
\usepackage{xcolor}
\usepackage{booktabs}
\usepackage{tabularx}
\usepackage{graphicx} 
\usepackage{lipsum}   
\usepackage{subcaption}
\usepackage{amsthm}
\glsdisablehyper

\newacronym{icl}{ICL}{In-context learning}
\newacronym{llm}{LLM}{large language model}
\newacronym{sc}{SC}{Surprise Calibration}

\definecolor{green}{RGB}{72,164,63}
\definecolor{red}{RGB}{193,18,28}

\title{{Surprise Calibration for Better In-Context Learning}}

\author{Zhihang Tan$^1$ \hspace{1em}
        Jingrui Hou$^1$ \hspace{1em}
        Ping Wang$^{1,2}$ \hspace{1em}
        Qibiao Hu$^1$ \hspace{1em}
        Peng Zhu$^3$ \\
        \normalsize $^1$School of Information Management, Wuhan University \\
        \normalsize $^2$Center for the Studies of Information Resources, Wuhan University \\
        \normalsize $^3$School of Economics and Management, Nanjing University of Science and Technology \\
        \texttt{\{zhihangtan, houjingrui, wangping, huqibiao\}@whu.edu.cn, pzhu@njust.edu.cn}}

\begin{document}
\maketitle 
\begin{abstract}
In-context learning (ICL) has emerged as a powerful paradigm for task adaptation in large language models (LLMs), where models infer underlying task structures from a few demonstrations. However, ICL remains susceptible to biases that arise from prior knowledge and contextual demonstrations, which can degrade the performance of LLMs. Existing bias calibration methods typically apply fixed class priors across all inputs, limiting their efficacy in dynamic ICL settings where the context for each query differs. To address these limitations, we adopt implicit sequential Bayesian inference as a framework for interpreting ICL, identify “surprise” as an informative signal for class prior shift, and introduce a novel method—Surprise Calibration (SC). SC leverages the notion of surprise to capture the temporal dynamics of class priors, providing a more adaptive and computationally efficient solution for in-context learning. We empirically demonstrate the superiority of SC over existing bias calibration techniques across a range of benchmark natural language processing tasks. 
\end{abstract}

\section{Introduction}

In recent years, \gls{icl} has emerged as a powerful paradigm to enable \glspl{llm} to adapt to natural language processing tasks with minimal supervision \citep{radford2019language,liu2023pre,dong2022survey,zhou2024mystery}. Unlike traditional methods that rely on retraining, \gls{icl} enables models to adapt to new tasks by simply providing a set of demonstrations within the input context. While \gls{icl} has demonstrated significant success, it remains susceptible to biases, a phenomenon reflecting \glspl{llm}' tendency to make predictions that favor certain categories over others \citep{min2022rethinking,holtzman2021surface,levine2021inductive,si2023measuring}. These biases can degrade the performance of the \glspl{llm}, For example, if a class prior overemphasizes one category, the model may be more likely to predict that category, even when it's not the most relevant or accurate choice.


\begin{figure}[t]
\resizebox{0.50\textwidth}{!}{
    \centering
        \includegraphics[width=0.8\textwidth]{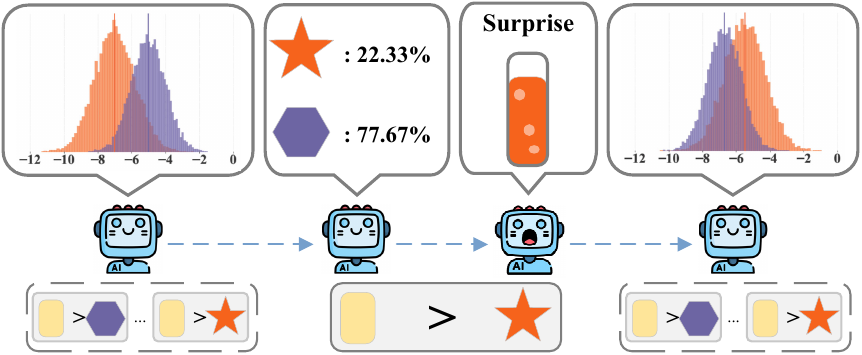}%
        }
    \caption{Illustration of surprise signal. The yellow rectangle denotes the example input in ICL, while the star and hexagon represent different class labels. The ">" symbol indicates the separator. A mismatch between the model's prediction and the ground-truth label leads to surprise, which in turn causes a shift in the class prior.}
    \label{fig:surprise}
\end{figure}
This issue has spurred research into developing methods to calibrate biases within \gls{icl} \citep{zhao2021calibrate,han2022prototypical,zhou2023batch,abbas2024enhancing}.  However, most existing methods perform a one-time calibration of class priors and apply them uniformly across all inputs. 
While such techniques outperform vanilla \gls{icl}, they fall short in dynamic \gls{icl} scenarios,  where each input instance may be paired with distinct contextual demonstrations \citep{rubin2022learning,zhang2022active,li2023finding,shu2024comparative,luo2024incontextlearningretrieveddemonstrations}. As a result, such techniques cannot adapt efficiently to the changing context.

In this work, we adopt implicit sequential Bayesian inference as a lens to interpret \gls{icl}, modeling it as a process in which \glspl{llm} continually update their implicit beliefs upon observing new demonstrations. Under this perspective, we propose the notion of \emph{surprise} as a potential signal for prior shift, which quantifies how ``surprised'' \glspl{llm} are when encountering new demonstrations(as shown in Figure~\ref{fig:surprise}). And our empirical analysis verifies that surprise functions as a robust and effective mechanism for class prior adjustment, even when other contextual biases are present.

 
Based on these insights, we propose a novel method: \gls{sc}. \Gls{sc} integrates surprise as a signal for detecting a shift in class prior. Rather than relying on fixed priors, \gls{sc} dynamically updates class priors based on the evolving context, allowing the model to adapt its predictions in the context of dynamic \gls{icl}. 

We evaluate the effectiveness of \gls{sc} across eight datasets from six NLP tasks. Our experiments demonstrate that \gls{sc}  outperforms state-of-the-art \gls{icl} bias calibration baselines and shows superior robustness and stability. 

The contributions of this paper are as follows:
\begin{itemize} 

\item We reinterpret in-context learning through the lens of implicit sequential Bayesian inference, and identify surprise as a core signal that drives class prior adjustment.

\item We introduce \glsdesc{sc}, a novel approach to dynamically calibrating \gls{icl} based on the cumulative effect of surprise signals on class priors. 

\item We empirically demonstrate the superior performance and computational efficiency of \gls{sc} on a range of NLP tasks, showing its potential for scalable and robust \gls{icl} in real-world applications.

\end{itemize}

\section{Background}  

\subsection{In-Context Learning}

\Gls{icl} refers to the ability of a \gls{llm}  to adapt to new tasks or instructions by leveraging a set of provided demonstrations (prompts) without requiring explicit retraining \citep{radford2019language,liu2023pre,dong2022survey}.  Typically, given a query (a test instance for which the model needs to generate a prediction) and context ( a set of demonstrations), \gls{icl} leverages the knowledge encoded in the pre-trained model parameters and the contextual information to generate an answer \citep{brown2020language}.

Formally, let $\mathcal{E}$ denote the space of all possible input queries, and let $\mathcal{Y}$ represent the space of all possible outputs (i.e., labels). For a given query $e \in \mathcal{E}$, the model produces a predicted label $\hat{y}$ by evaluating the conditional probability distribution over all candidate labels $y' \in \mathcal{Y}$: 
    \begin{equation}
    \text{\small$ \hat{y} = \arg \max_{y' \in \mathcal{Y}} p(y' \mid e, D),$} \end{equation} 
where $D$ represents the context. Early implementations of \gls{icl} relied on a fixed set of demonstrations (either hand-crafted or randomly selected) for all potential queries \citep{hendrycksmath2021,wei2022chain,lewkowycz2022solving}. In such setups, the context $D$ consists of $K$ examples, i.e., $D = \{(e_j, y_j\}_{j=1}^K$, where each pair $(e_j, y_j)$ corresponds to an example, with $e_j \in \mathcal{E}$ and $y_j \in \mathcal{Y}$.

However, the performance of \gls{icl} is sensitive to the selection of demonstrations\citep{liu-etal-2022-makes,qin2023context}. Recent advancements have introduced methods for dynamically selecting query-specific demonstrations, which are collectively referred to as \emph{dynamic \gls{icl}} \citep{qin2023context,wang2024large,peng2024revisiting}. Unlike static demonstrations, dynamic \gls{icl} tailors the context $D$ for each individual query $e$. One typical approach involves selecting the most relevant $K$ examples to the query, optimizing model performance based on factors such as semantic similarity or task-specific relevance \citep{liu-etal-2022-makes, luo2024incontextlearningretrieveddemonstrations}.

\subsection{Bias of In-Context Learning}

Despite its empirical success in few-shot learning, \gls{icl} has been shown to be susceptible to systematic biases that can distort predictions. \citet{zhao2021calibrate} highlighted that the biases observed in \glspl{llm} predominantly stem from two key sources: the contextual demonstrations provided and the inherent characteristics of the model itself. Contextual bias encompasses issues such as majority label bias, where models tend to favor the most frequent label in the demonstration set, and recency label bias, where models show a preference for labels that appear at the end of the demonstration sequence. On the other hand, model-intrinsic bias reflects the model's inherent propensity to predict label tokens that were frequent in its pretraining corpus. These biases can significantly undermine the robustness and reliability of \gls{icl} methods, leading to suboptimal performance in real-world applications.

To mitigate these biases, prior work \citep{zhao2021calibrate,han2022prototypical,fei2023mitigating,zhou2023batch,abbas2024enhancing} has suggested calibrating the predicted probabilities of \glspl{llm} by incorporating class prior probabilities, which can be estimated through repeated sampling. Formally, the prior probability for any label $y' \in \mathcal{Y}$ can then be expressed as: 
    \begin{equation}
    \text{\small$
p(y') = p(y' \mid D).$}
\end{equation} 

In practice, estimating this prior probability typically involves repeated sampling. For a fixed context $D$, the sampling process is repeated $n$ times (e.g., with a fixed value $n=3$, as used in \citet{zhao2021calibrate}). However, explicitly accounting for contextual bias by performing separate prior estimation for each query can lead to prohibitively high computational costs. Conversely, disregarding contextual bias entirely may result in suboptimal performance, as it fails to capture critical variations in the prior distribution. This trade-off is particularly pronounced in \emph{dynamic \gls{icl}} settings, where the context $D$ varies for each query, making repeated sampling impractical for real-time or resource-constrained applications.

This work seeks to address the aforementioned challenge by proposing an efficient method for estimating class priors in dynamic \gls{icl} scenarios. Unlike static \gls{icl}, accurately estimating the prior in dynamic settings requires a comprehensive understanding of the process by which \gls{icl} models utilize demonstrations for inference. To this end, we first establish a theoretical framework to model this process and subsequently introduce a novel approach that enhances the accuracy and robustness of dynamic \gls{icl}.

\section{Surprise: a Signal for Class Prior Shift}
\label{sec:surprise-prior modeling}
\begin{figure*}[htb]
    \centering
    \resizebox{\textwidth}{!}{
        \includegraphics[width=0.25\textwidth]{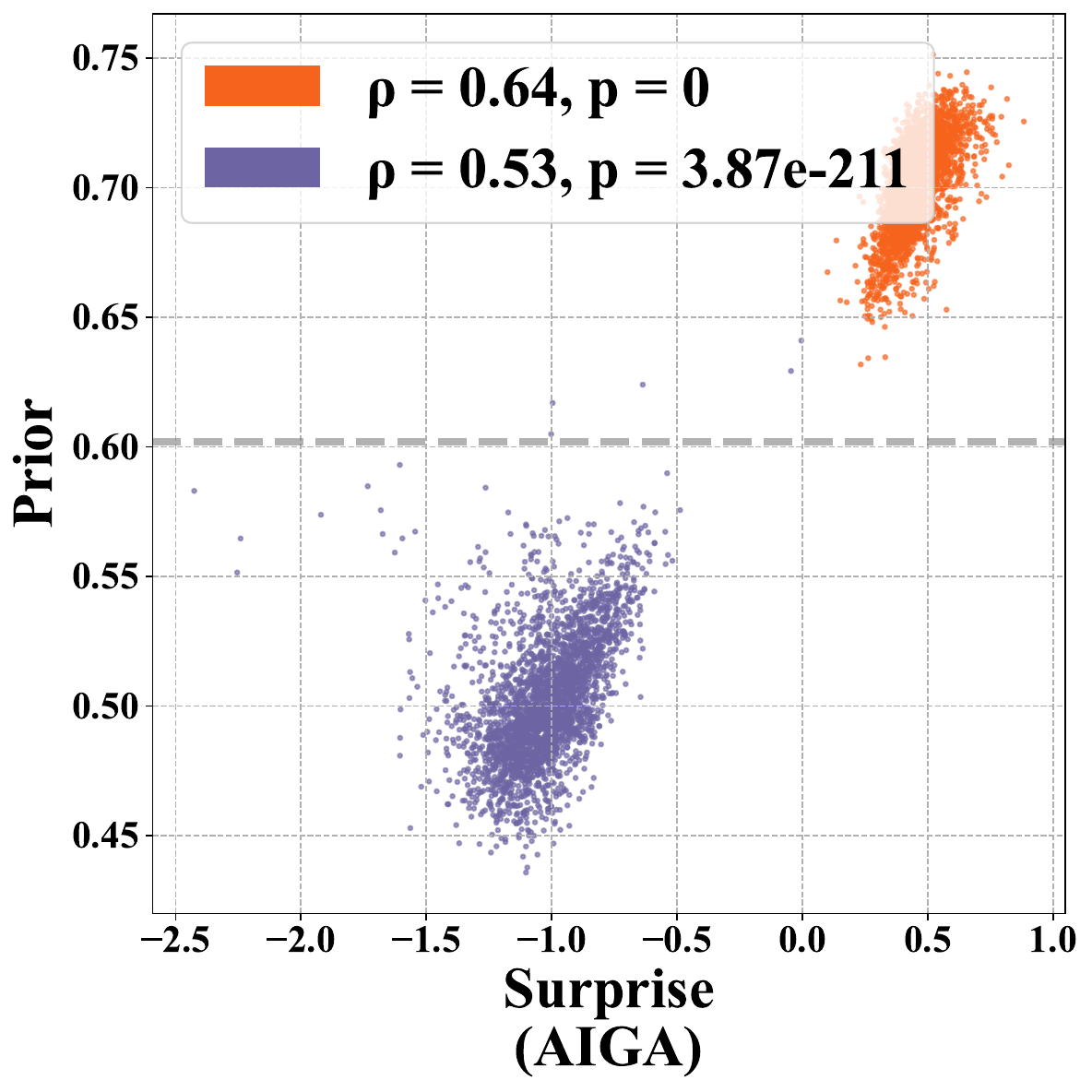}%
        \includegraphics[width=0.25\textwidth]{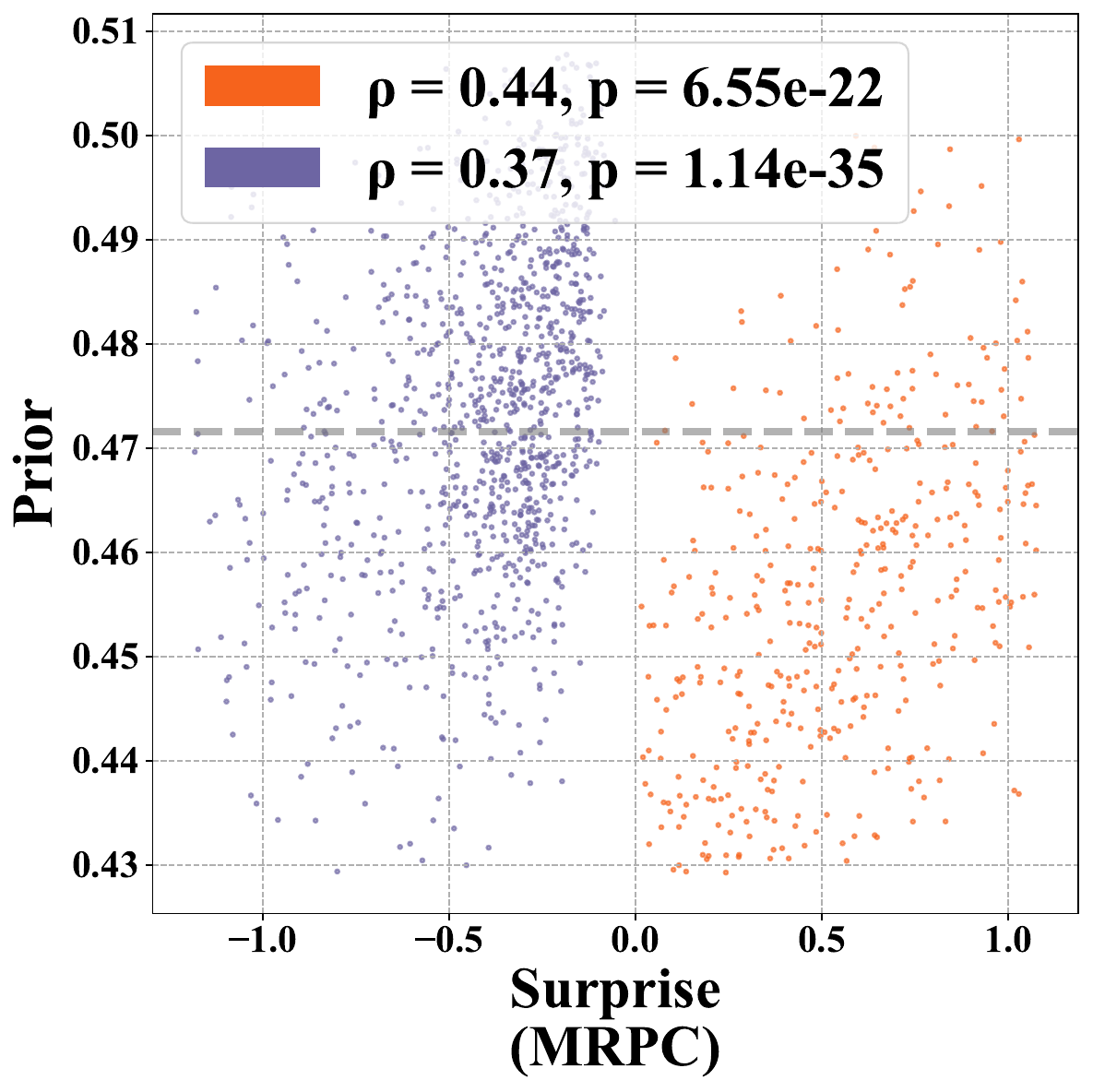}%
        \includegraphics[width=0.25\textwidth]{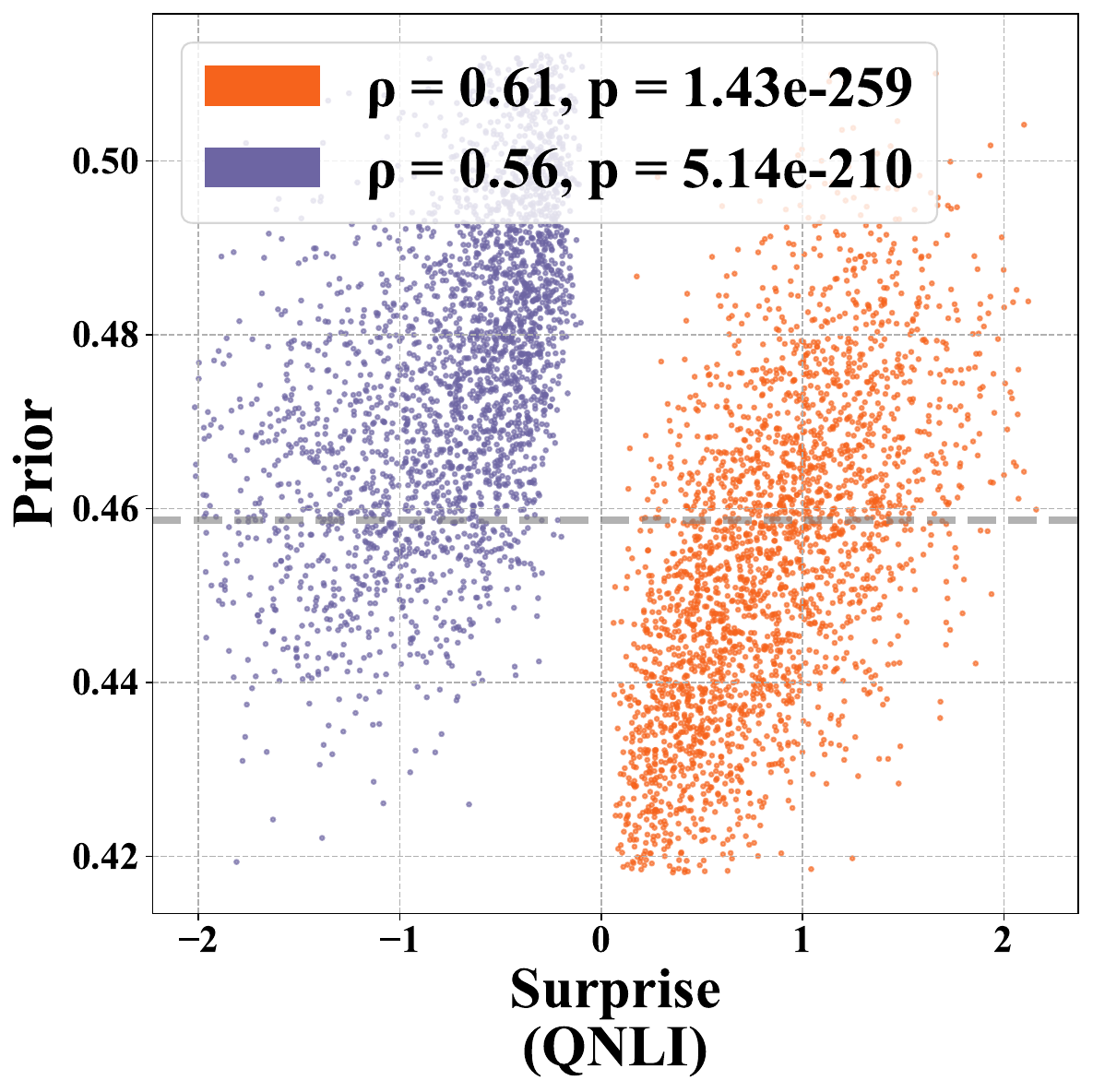}%
        \includegraphics[width=0.25\textwidth]{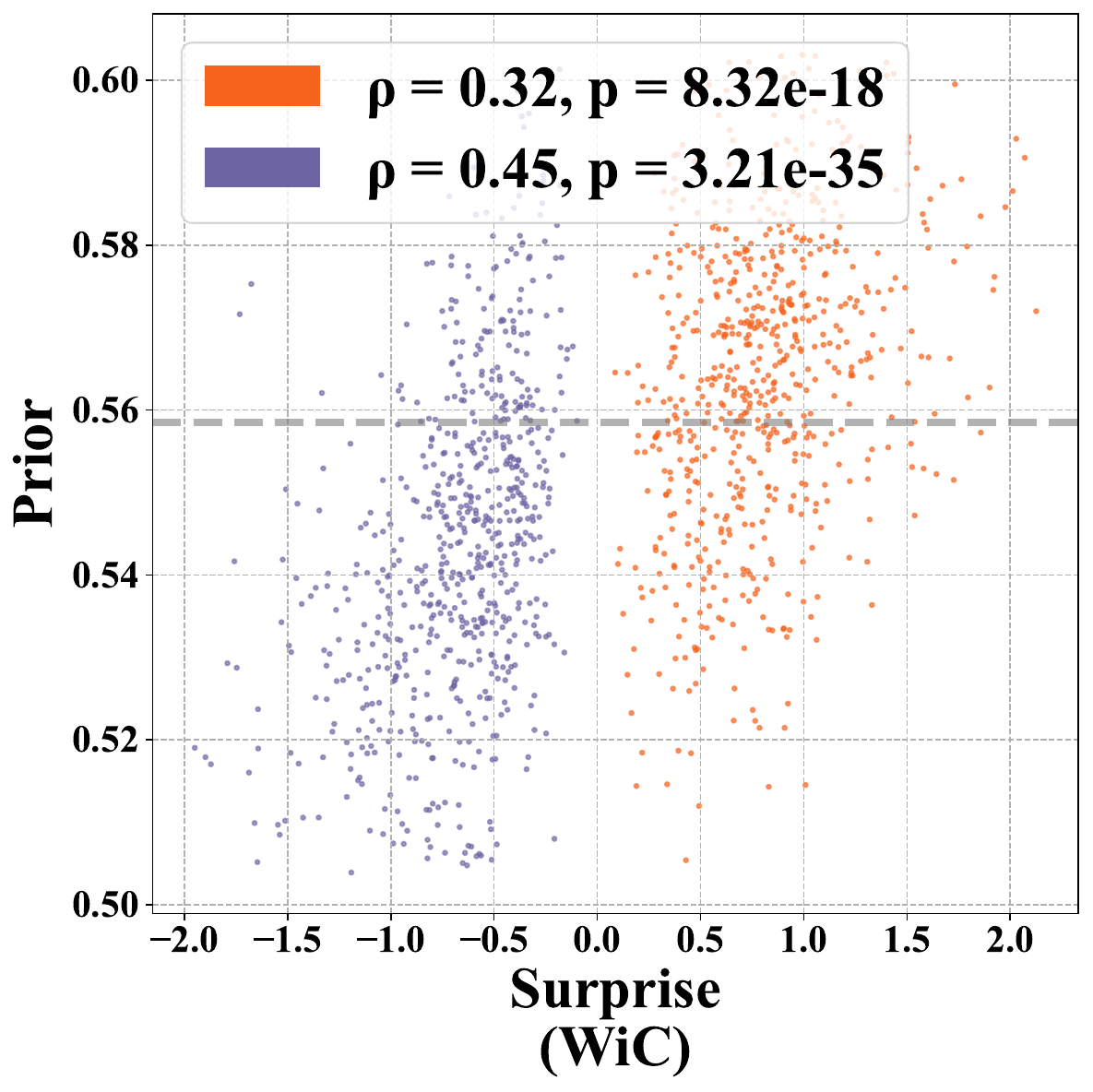}%
    }
    \caption{Spearman correlation between surprise ($-\log p(y|e, D)$) and the prior probability of the positive class across four binary classification datasets (AIGA, MRPC, QNLI, WiC). Each scatter plot shows the relationship of surprise and positive class prior for positive (orange) and negative (purple) demonstration insertions, with corresponding Spearman $\rho$ and $p$-values.
    The gray dashed line marks the estimated prior before insertion. All priors are estimated by repeated sampling using the BC method (described in Section \ref{sec:Experimental Setup}).}
    \label{fig:surprise-validation}
\end{figure*}

\subsection{Reinterpretation ICL via Implicit Sequential Bayesian Inference}
\label{sec:seq-updates}

We first highlight two essential characteristics of \gls{llm} that motivate our interpretation of in-context learning as implicit sequential Bayesian inference.

 \noindent \textbf{(1) Latent concept.} In the context of \gls{icl}, prior works have proposed that transformers implicitly infer a latent concept \( z \) underlying the demonstration data \( D \), which is subsequently leveraged to guide predictions \citep{xie2021explanation, tefnik2023can, hendel2023context, han2024emergence, todd2024function}. Here, $z \in \mathcal{Z}$, where $\mathcal{Z}$ denotes the space of all possible concepts that explain the demonstrations, such as label assignment rules, task definitions, or underlying generative functions, depending on the task. Given this latent representation, the probability of predicting a label  for a new input  is modeled by marginalizing over all possible concepts:
    \begin{equation}
    \text{\small$
    p(y \mid e, D) = \int_{\mathcal{Z}} p(y \mid e, z) \, p(z \mid D) \, dz.$}
    \end{equation}

 \noindent \textbf{(2) Autoregressive language models and surprise signals.} Large language models (\glspl{llm}) generate output autoregressively, one token at a time \citep{kossen2024context}. Given an input sequence $(X_1, \dots, X_M)$, each token $X_i$ is predicted based on its preceding context $(X_1, \dots, X_{i-1})$. In \gls{icl}, prompts often structure each demonstration as: input $e$, followed by a label delimiter (e.g., `>' or `:'), then the corresponding label $y$. As the model processes these demonstrations sequentially, the hidden state at the delimiter captures its current belief about the label for $e_j$, conditioned on all previous $j-1$ demonstrations $D_{j-1}$—formally denoted as $p(y' \mid e_j, D_{j-1})$.
    
Crucially, this architecture allows us to measure the model’s predictive uncertainty before observing each label. Specifically, the conditional probability $p(y_j \mid e_j, D_{j-1})$, computed just before the true label $y_j$ is revealed, reflects how "surprised" the model is by the label. We define \textit{surprise} as the negative log probability of the observed label:  
    \begin{equation}
    \text{\small$
     \text{Surprise}(y_j \mid e_j, D_{j-1}) = -\log p(y_j \mid e_j, D_{j-1}).$}
     \end{equation}
     Higher values indicate greater surprise, signaling a mismatch between the model’s expectations and the observed label.

\textbf{Surprise-driven bayesian updating.} Building upon this intuition, we formalize the implicit Bayesian updating procedure based on observed demonstrations.
Formally, let the model’s belief over $z$ prior to observing a new example be $p(z \mid D_{j-1})$. Incorporating $(e_j, y_j)$ leads to a Bayesian update:
\begin{equation}
\text{\small$
p(z \mid D_{j-1} \cup \{(e_j, y_j)\}) = \frac{p(e_j, y_j \mid z)\, p(z \mid D_{j-1})}{\int_{\mathcal{Z}} p(e_j, y_j \mid z)\, p(z \mid D_{j-1})\, dz}.$}
\end{equation}

Given this update, the revised class prior for an arbitrary label $y'$ becomes:
\begin{align}
\label{eq:updated_classprior}
\text{\small$ p(y' \mid D_{j-1}$} & \text{\small$\cup \{(e_j, y_j)\}) =  $} 
\text{\small$\mathbb{E}_{z\sim p(z\mid D_{j-1})}[p(y'\mid z)]$}\\  \nonumber
&\text{\small$
+\frac{\mathrm{Cov}_{z\sim p(z\mid D_{j-1})}(p(y'\mid z),\,p(e_j,y_j\mid z))}{\mathbb{E}_{z\sim p(z\mid D_{j-1})}[p(e_j,y_j\mid z)]}
$}
\end{align}
Here, the denominator 
represents the expected joint likelihood of $(e_j, y_j)$ under the current belief over $z$. Comparing this to the predictive probability:
\begin{align}
\label{eq:pred_prob_decompose}
\text{\small$p(y_j \mid e_j, D_{j-1}) $}
&\text{\small$= \mathbb{E}_{z \sim p(z \mid D_{j-1})} \left[ p(y_j \mid e_j, z) \right]$} \nonumber \\
&\text{\small$= \mathbb{E}_{z \sim p(z \mid D_{j-1})} \left[ \frac{p(e_j, y_j \mid z)}{p(e_j \mid z)} \right].$}
\end{align}
We note that a low value of $p(y_j \mid e_j, D_{j-1})$ may suggest lower $\mathbb{E}_{z\sim p(z\mid D_{j-1})}[p(e_j,y_j \mid z)]$, though this relationship is not formally guaranteed due to the normalization by $p(e_j\mid D_{j-1})$. Nevertheless, we hypothesize that surprise—captured via $p(y_j \mid e_j, D_{j-1})$—can serve as a practical and meaningful signal for class prior adjustment. In particular, higher surprise(smaller value of $p(y_j\mid e_j,D_{j-1})$) is expected to amplify the influence of the covariance term, leading to larger shift in the predicted distribution over labels.

In summary, we highlight two possible mechanisms driving prior updates:
\begin{itemize}
\item  \textbf{Covariance-driven adjustment:} Positive covariance between $p(y'\mid z)$ and $p(e_j,y_j\mid z)$ (i.e. $ y'= y_j$) increases the posterior probability of $y'$, while negative covariance(i.e. $ y'\neq y_j$) decreases it (See Eq~\eqref{eq:updated_classprior}).
\item  \textbf{Surprise Amplification via Joint Likelihood:} Higher surprise (corresponding to a lower average joint likelihood) enhances the sensitivity of the update to variations in $z$, thereby amplifying belief shifts (See Eq~\eqref{eq:updated_classprior} and Eq~\eqref{eq:pred_prob_decompose}).
\end{itemize}

\subsection{Empirical Evidence}

To empirically validate that \textbf{surprise serves as a practical signal of prior shift}, we conducted controlled experiments on four binary classification datasets. For each dataset, we specifically focused on observing changes in the prior probability of a designated positive class upon inserting new examples.

By introducing a new example $(e, y)$ into a fixed set of demonstrations, we recorded the following information:
\begin{itemize}
\item the prior probability of the positive class both before and after the insertion;
\item the surprise value associated with the example, quantified as $-\log p(y \mid e, z)$.
\end{itemize}
Following the covariance-driven Bayesian updating framework, surprise values were assigned negative signs for negative-class examples and positive signs for positive-class examples, reflecting their respective suppressive and reinforcing effects on the positive class prior.
By repeating this process with different examples added to the fixed context, we collected multiple data points, each consisting of the surprise value and its corresponding positive class prior.

Figure~\ref{fig:surprise-validation} presents the relationship between surprise and prior shift across datasets. Crucially, across all experimental setups, we consistently observed statistically significant positive Spearman correlations between surprise and the shift in the positive-class prior. Moreover, we observe that as the absolute value of surprise for positive-class examples increases, the class prior shifts toward the positive class. Conversely, as the absolute value of surprise for negative-class examples increases, the class prior shifts toward the negative class (manifested in the figure as a decrease in the positive-class prior).  This result robustly supports our theoretical assertion that surprise systematically co-varies with prior adjustments, effectively predicting the direction and magnitude of these shifts.

Importantly, this correlation was most pronounced within groups conditioned on the newly inserted example’s label. Aggregate correlations occasionally showed instability, which we attribute to an additional influence—\textit{anti-recency bias}. This bias tends to shift overall predictions away from recently observed labels, creating an offset in mean prior probabilities between groups. We further investigate the nature of this bias in Figure \ref{fig:anti-recency bias} in the Appendix.

Despite the presence of anti-recency bias, our findings emphasize a critical distinction: while biases may alter the overall mean prior levels, the structural relationship between surprise and prior adjustment remains robust and monotonic. Surprise reliably signals local updates to the model's beliefs about class prior.
\section{Surprise Calibration}
\label{sec:surprise_sequence}
\begin{figure}[t]
\resizebox{0.45\textwidth}{!}{
    \centering
        \includegraphics[width=0.8\textwidth]{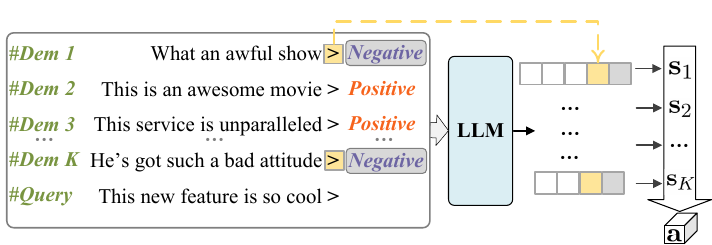}%
        }
    \caption{Illustration of the proposed \glsdesc{sc} for \glsdesc{icl}.}
    \label{fig:method}
\end{figure}
Building on the insights from Section~\ref{sec:surprise-prior modeling}, we propose \textbf{\gls{sc}}, a novel calibration method that explicitly models the temporal propagation of surprise signals during \gls{icl} as illustrated in Figure~\ref{fig:method}.

To operationalize surprise-driven calibration, we begin by constructing a \textbf{surprise vector}  for each in-context demonstration. Given a context example $(e_j, y_j)$ and the model’s current belief (conditioned on prior context $D_{j-1}$), we define the surprise vector $\mathbf{s} \in \mathbb{R}^C$ ($C$ = number of classes), where the $c$-th dimension is
\begin{equation}
\text{\small$
s_c = \bigl(1 - 2\,\delta_{c,y}\bigr)\,\log p\bigl(y = c \mid e_j,\,D_{j-1}\bigr).
$}\end{equation}
where
\begin{equation}
\text{\small$
\delta_{c,y} =
\begin{cases}
1, & c = y,\\
0, & c \neq y. \nonumber
\end{cases}
$}\end{equation}
Here, the sign of $s_c$ indicates whether class $c$ matches the true label ($c = y$: negative, $c \ne y$: positive), and the magnitude reflects how surprising the label is to the model. The sign encodes the expected direction of prior adjustment for that class.

By collecting surprise vectors for each context example, we form a \textbf{surprise sequence}, i.e., $\mathcal{S} = [\mathbf{s}_1, \ldots, \mathbf{s}_K]$, which explicitly represents both the amount and direction in which each demonstration should modulate the model’s evolving class prior.

To capture dependencies across this sequence and aggregate the adjustment signals, we use a time series prediction model (e.g. GRU) to process the surprise sequence. The final hidden state of the GRU is decoded into a \textbf{prior adjustment vector} $\mathbf{a} \in \mathbb{R}^C$, which models the cumulative shift in class priors based on the observed context.

For a query input $e$, we compute the model’s original prediction probability over label $y'$: $p_\text{orig}(y' \mid e, z)$. Our calibrated probability is then obtained by adjusting the original log-probability with the corresponding prior adjustment:
\begin{equation}
\text{\small$
-\log p_\text{calib}(y' \mid e, z) = -\log p_\text{orig}(y' \mid e, z) - a_{y'}
$}\end{equation}
where $a_{y'}$ is the adjustment for the class $y'$ inferred from the surprise sequence. This effectively shifts the model’s confidence in the predicted class according to the accumulated surprise from previous in-context demonstrations.

The model is trained end-to-end by minimizing the cross-entropy loss between the calibrated probabilities and the true label of the query input, jointly optimizing the parameters of the time series prediction model and decoding layers.

\section{Experiments}
\subsection{Experimental Setup}
\label{sec:Experimental Setup}
\paragraph{Datasets}
We evaluate the effectiveness of our \gls{sc} method on 8  datasets within 6 natural language tasks. Specifically, we consider sentiment classification: SST-2 \citep{socher2013recursive}; natural language inference and entailment: RTE, QNLI, MNLI \citep{williams2018broad}; paraphrasing: MRPC \citep{dolan2005automatically}; word disambiguation: WiC \citep{pilehvar2019wic}; spam detection: YouTube Spam(YouTube) \citep{alberto2015tubespam}; AIGC detection: AIGA \citep{10233982}. 
\paragraph{Models}
We conducted experiments on two pre-trained models with different sizes from the Qwen2.5 series \citep{qwen2.5}, Qwen2.5-3B, and Qwen2.5-7B. The Qwen2.5 series has shown highly competitive performance, matching or surpassing other models with similar parameter counts.

\paragraph{Baselines}
We compare our method with five advanced calibration methods: \textbf{ICL}: Vanilla in-context learning performance; \textbf{BC} (Batch Calibration) \citep{zhou2023batch}: Estimates class priors directly from a batch of input data; \textbf{LinC} (Linear Probe Calibration) \citep{abbas2024enhancing}: Estimates class priors using a labeled training dataset; \textbf{CC+} (Contextual Calibration) \citep{zhao2021calibrate}: Estimates class priors per query via content-free examples; \textbf{BC+} and \textbf{LinC+}: Query-specific extensions of BC and LinC, respectively.
The ``+'' symbol indicates that class priors are estimated individually for each query, which introduces additional inference overhead as shown in Table~\ref{tab:inference_times}. 
\begin{table}[htbp]
\centering
\resizebox{0.25\textwidth}{!}{
\begin{tabular}{lc}
\toprule
Method & Inference Count \\
\midrule
BC     & $T$               \\
LinC   & $M+T$               \\
CC+    & $3×T$             \\
BC+    & $n×T$             \\
LinC+  & $n×T$             \\
Ours   & $M+T$             \\
\bottomrule
\end{tabular}
}
\caption{Comparison of Inference Count for Different Methods. $T$ represents the number of samples to be predicted, $n$ denotes the additional sample size used for estimating class priors in BC+ and LinC+, and $M$ is the sample size used for training the SC and LinC model.}
\label{tab:inference_times}
\end{table}
\paragraph{Demonstrations Selecting and Ordering Strategies} 
We explore three demonstration selection strategies: \textbf{Random}, which randomly selects contextual examples; \textbf{BM25}, which uses word-overlap similarity to choose high-similarity demonstrations; and \textbf{Top-k}, which selects nearest neighbors based on cosine similarity of embeddings generated by the GTE model\citep{li2023towards}. For BM25 and Top-k, we apply three ordering strategies: \textbf{Increase}, placing higher-value examples at the end; \textbf{Decrease}, placing them at the beginning; and \textbf{U-curve}, positioning higher-value examples at both ends and lower-value ones in the middle.

\subsection{Main Experimental}
\begin{table*}[t]
\centering
\resizebox{\textwidth}{!}{
\begin{tabular}{@{\extracolsep{\fill}} l c c c c c c c c}
\toprule
\multirow{2}{*}{\textbf{DataSet}} & \multirow{2}{*}{\textbf{LM}} & \multicolumn{7}{c}{\textbf{Method}} \\
 & &  \textbf{ICL} & \textbf{BC} & \textbf{LinC} & \textbf{CC+} & \textbf{BC+} & \textbf{LinC+} & \textbf{Ours} \\
\midrule
SST-2 & Qwen2.5-3B & 89.68\% & 89.79\% \textcolor{green}{(+0.11\%)} & 89.90\% \textcolor{green}{(+0.22\%)} & 88.58\% \textcolor{red}{(-1.10\%)} & 89.84\% \textcolor{green}{(+0.16\%)} & \textbf{90.33\%} \textcolor{green}{(+0.65\%)} & 89.68\%\textcolor{green}{(+0.00\%)}\\
& Qwen2.5-7B & 94.45\% & 94.51\% \textcolor{green}{(+0.06\%)} & 94.51\% \textcolor{green}{(+0.06\%)} & \textbf{95.00\%} \textcolor{green}{(+0.55\%)} & 94.50\% \textcolor{green}{(+0.05\%)} & 94.67\% \textcolor{green}{(+0.22\%)} & \textbf{95.00\%}\textcolor{green}{(+0.55\%)}\\
\midrule
MNLI & Qwen2.5-3B & 66.77\% & 72.38\% \textcolor{green}{(+5.61\%)} & 73.44\% \textcolor{green}{(+6.67\%)} & 70.18\% \textcolor{green}{(+3.41\%)} & \textbf{79.07\%} \textcolor{green}{(+12.30\%)} & 67.52\% \textcolor{green}{(+0.75\%)} & 76.33\% \textcolor{green}{(+9.56\%)} \\
& Qwen2.5-7B & 67.01\% & 75.03\% \textcolor{green}{(+8.02\%)} & 77.27\% \textcolor{green}{(+10.26\%)} & 79.78\% \textcolor{green}{(+12.77\%)} & 75.02\% \textcolor{green}{(+8.01\%)} & 67.60\% \textcolor{green}{(+0.59\%)} & \textbf{81.15\%} \textcolor{green}{(+14.14\%)} \\
\midrule
MRPC & Qwen2.5-3B & 68.00\% & 67.42\% \textcolor{red}{(-0.58\%)} & 67.77\% \textcolor{red}{(-0.23\%)} & 71.71\% \textcolor{green}{(+3.71\%)} & 71.71\% \textcolor{green}{(+3.71\%)} & 70.66\% \textcolor{green}{(+2.66\%)} & \textbf{72.28\%} \textcolor{green}{(+4.28\%)}\\
& Qwen2.5-7B & 68.17\% & 67.82\% \textcolor{red}{(-0.35\%)} & 72.35\% \textcolor{green}{(+4.18\%)} & 73.33\% \textcolor{green}{(+5.16\%)} & 73.97\% \textcolor{green}{(+5.80\%)} & 72.86\% \textcolor{green}{(+4.69\%)} & \textbf{73.85\%} \textcolor{green}{(+5.68\%)}\\
\midrule
QNLI & Qwen2.5-3B & 72.67\% & 76.57\% \textcolor{green}{(+3.90\%)} & 76.51\% \textcolor{green}{(+3.84\%)} & 69.47\% \textcolor{red}{(-3.20\%)} & \textbf{79.53\%} \textcolor{green}{(+6.86\%)} & 79.26\% \textcolor{green}{(+6.59\%)} & 78.01\% \textcolor{green}{(+5.34\%)}\\
& Qwen2.5-7B & 80.26\% & 80.34\% \textcolor{green}{(+0.08\%)} & 80.19\% \textcolor{red}{(-0.07\%)} & 74.48\% \textcolor{red}{(-5.78\%)} & 80.83\% \textcolor{green}{(+0.57\%)} & \textbf{81.01\%} \textcolor{green}{(+0.75\%)} & 80.48\% \textcolor{green}{(+0.14\%)}\\
\midrule
RTE & Qwen2.5-3B & 75.81\% & 82.67\% \textcolor{green}{(+6.86\%)} & 83.03\% \textcolor{green}{(+7.22\%)} & 72.93\% \textcolor{red}{(-2.88\%)} & 75.45\% \textcolor{red}{(-0.36\%)} & 75.81\% \textcolor{green}{(+0.00\%)} & \textbf{84.47\%} \textcolor{green}{(+8.66\%)}\\
& Qwen2.5-7B & 83.39\% & \textbf{83.75\%} \textcolor{green}{(+0.36\%)} & \textbf{83.75\%} \textcolor{green}{(+0.36\%)} & 79.06\% \textcolor{red}{(-4.33\%)} & 81.22\% \textcolor{red}{(-2.17\%)} & 83.03\% \textcolor{red}{(-0.36\%)} & \textbf{83.75\%} \textcolor{green}{(+0.36\%)}\\
\midrule
WiC & Qwen2.5-3B & 57.07\% & \textbf{59.57\%} \textcolor{green}{(+2.50\%)} & 58.93\% \textcolor{green}{(+1.86\%)} & 54.93\% \textcolor{red}{(-2.14\%)} & \textbf{59.57\%} \textcolor{green}{(+2.50\%)} & 58.71\% \textcolor{green}{(+1.64\%)} & 58.29\% \textcolor{green}{(+1.22\%)} \\
& Qwen2.5-7B & 62.86\%& 63.50\% \textcolor{green}{(+0.64\%)} & 62.93\% \textcolor{red}{(-0.07\%)} & 60.57\% \textcolor{red}{(-2.29\%)} & 63.21\% \textcolor{green}{(+0.35\%)} & \textbf{63.92\%} \textcolor{green}{(+1.06\%)} & 62.93\% \textcolor{green}{(+0.07\%)}\\
\midrule
YouTube & Qwen2.5-3B & 86.48\% & 86.73\% \textcolor{green}{(+0.25\%)} & 87.24\% \textcolor{green}{(+0.76\%)} & 78.83\% \textcolor{red}{(-7.65\%)} & 77.04\% \textcolor{red}{(-9.44\%)} & 88.52\% \textcolor{green}{(+2.04\%)} & \textbf{90.56\%} \textcolor{green}{(+4.08\%)}\\
& Qwen2.5-7B & 87.50\% & 88.52\% \textcolor{green}{(+1.02\%)} & 88.52\% \textcolor{green}{(+1.02\%)} & 84.18\% \textcolor{red}{(-3.32\%)} & 80.87\% \textcolor{red}{(-6.63\%)} & 88.27\% \textcolor{green}{(+0.77\%)} & \textbf{90.56\%} \textcolor{green}{(+3.06\%)}\\
\midrule
AIGA & Qwen2.5-3B & 76.37\% & 76.97\% \textcolor{green}{(+0.60\%)} & 77.33\% \textcolor{green}{(+0.96\%)} & 75.91\% \textcolor{red}{(-0.46\%)} & 75.47\% \textcolor{red}{(-0.90\%)} & 77.12\% \textcolor{green}{(+0.75\%)} & \textbf{79.99\%} \textcolor{green}{(+3.62\%)}\\
& Qwen2.5-7B & 75.71\% & 77.20\% \textcolor{green}{(+1.49\%)} & 77.40\% \textcolor{green}{(+1.69\%)} & 79.16\% \textcolor{green}{(+3.45\%)} & 80.93\% \textcolor{green}{(+5.22\%)} & 82.43\% \textcolor{green}{(+6.72\%)} & \textbf{79.97\%} \textcolor{green}{(+4.26\%)}\\
\midrule
Avg. & Qwen2.5-3B & 74.11\% & 76.51\% \textcolor{green}{(+2.40\%)} & 76.77\% \textcolor{green}{(+2.66\%)} & 72.82\% \textcolor{red}{(-1.29\%)} & 75.96\% \textcolor{green}{(+1.85\%)} & 75.99\% \textcolor{green}{(+1.88\%)} & \textbf{78.70\%} \textcolor{green}{(+4.59\%)}\\
& Qwen2.5-7B & 77.42\% & 78.83\% \textcolor{green}{(+1.41\%)} & 79.62\% \textcolor{green}{(+2.20\%)} & 78.20\% \textcolor{green}{(+0.78\%)} & 78.82\% \textcolor{green}{(+1.40\%)} & 79.22\% \textcolor{green}{(+1.80\%)} & \textbf{80.96\%} \textcolor{green}{(+3.54\%)}\\
\bottomrule
\end{tabular}
}
\caption{Accuracy(\%) comparison of different calibration methods on various datasets using BM25 selecting strategy, increase ordering strategy and 3-shot Qwen2.5 models (3B and 7B). We train a GRU model as the backbone of SC framework and report results using fixed random seeds and hyper-parameters. The best performance for each dataset and model size is highlighted in bold. The percentage changes relative to ICL are shown in parentheses. Green indicates improvement, and red indicates a reduction.}
\label{tab:main_table}
\end{table*}
\begin{figure}[ht]
\centering
\begin{subfigure}{.23\textwidth}
\centering
\includegraphics[width=\textwidth]{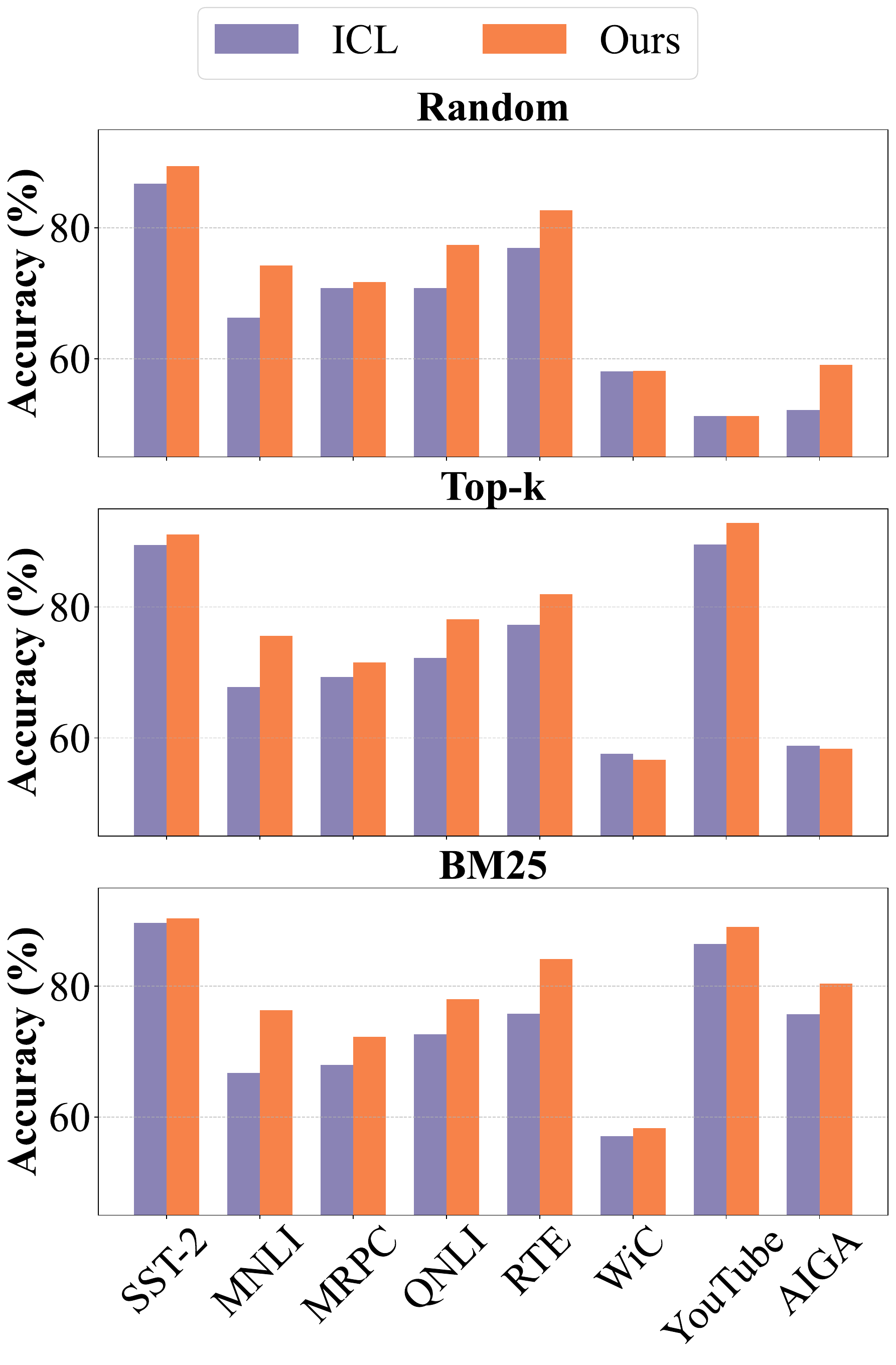}
\caption{3 demonstration selection.}

\end{subfigure}
\begin{subfigure}{.23\textwidth}
        \centering
        \includegraphics[width=\textwidth]{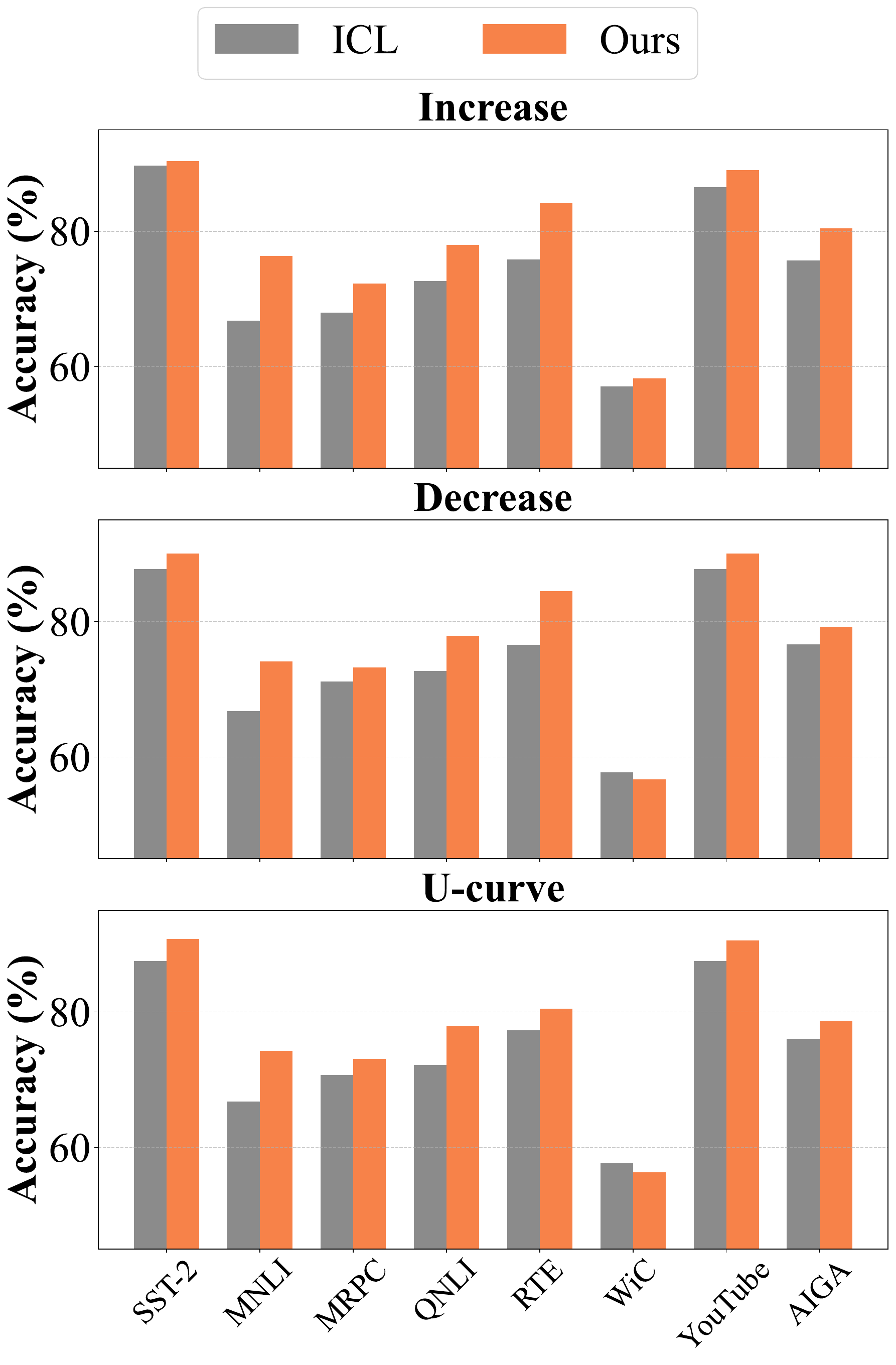}
        \caption{3 demonstration ordering.}

\end{subfigure}
\caption{Accuracy comparison between \gls{sc} and ICL across 3 demonstration selection and demonstration ordering strategies with other settings kept consistent as shown in Table \ref{tab:main_table}.}
\label{fig:selection}
\end{figure}
Table~\ref{tab:main_table} presents three main findings. Firstly, the proposed method, SC, consistently exhibits high performance across various model sizes and datasets. Specifically, SC outperforms ICL by +4.59\% on Qwen2.5-3B and +3.54\% on Qwen2.5-7B on average and outperforms or matches the best-performing baseline in most cases. Secondly, in contrast to other dynamic prior estimation techniques like CC+ and BC+, SC consistently outperforms ICL without any performance degradation in all settings. Lastly, SC achieves these results while maintaining computational efficiency, as it does not require additional multiplicative inference iterations to estimate class priors for each input query. Instead, it only requires a limited number of training samples to calibrate the priors, as Table \ref{tab:inference_times} shows.
\begin{figure}[t]
    \centering
        \includegraphics[width=0.25\textwidth]{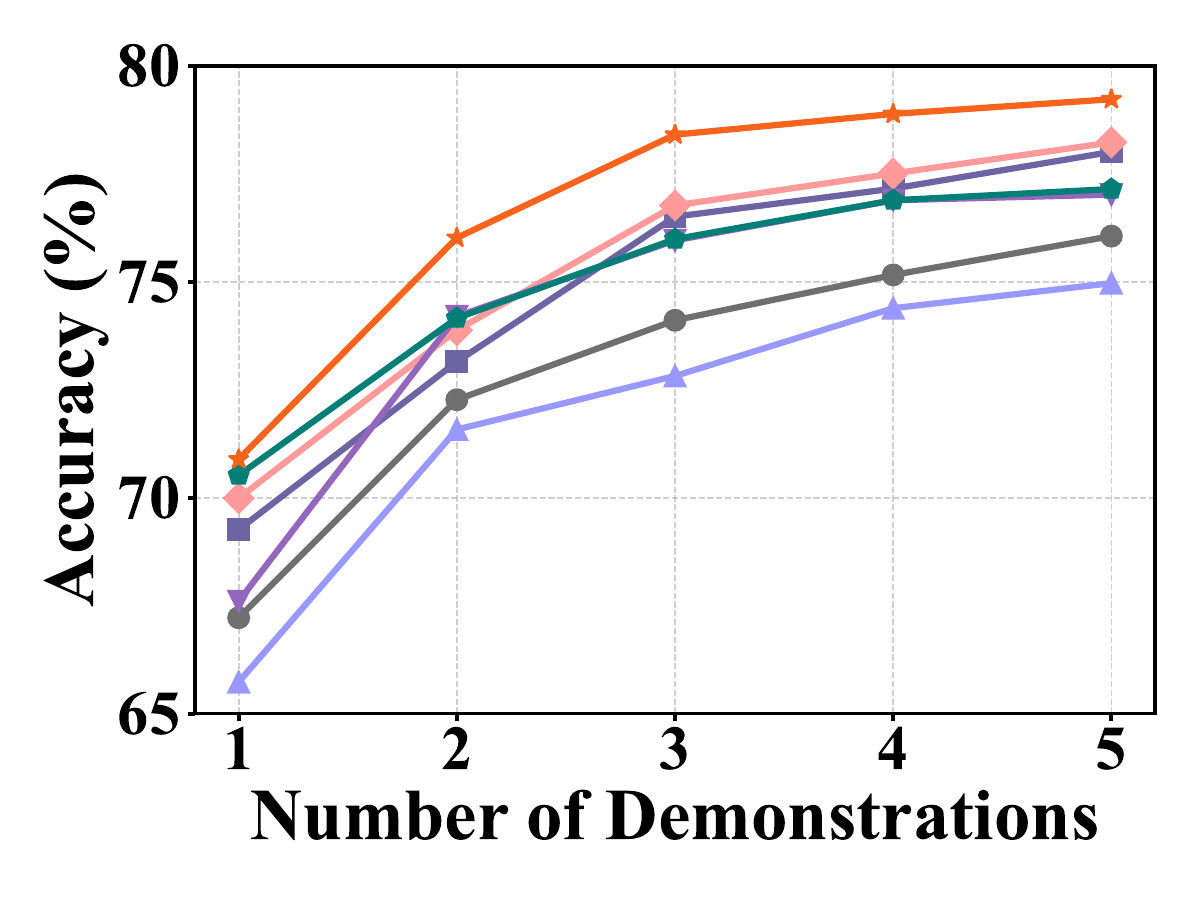}%
        \includegraphics[width=0.25\textwidth]{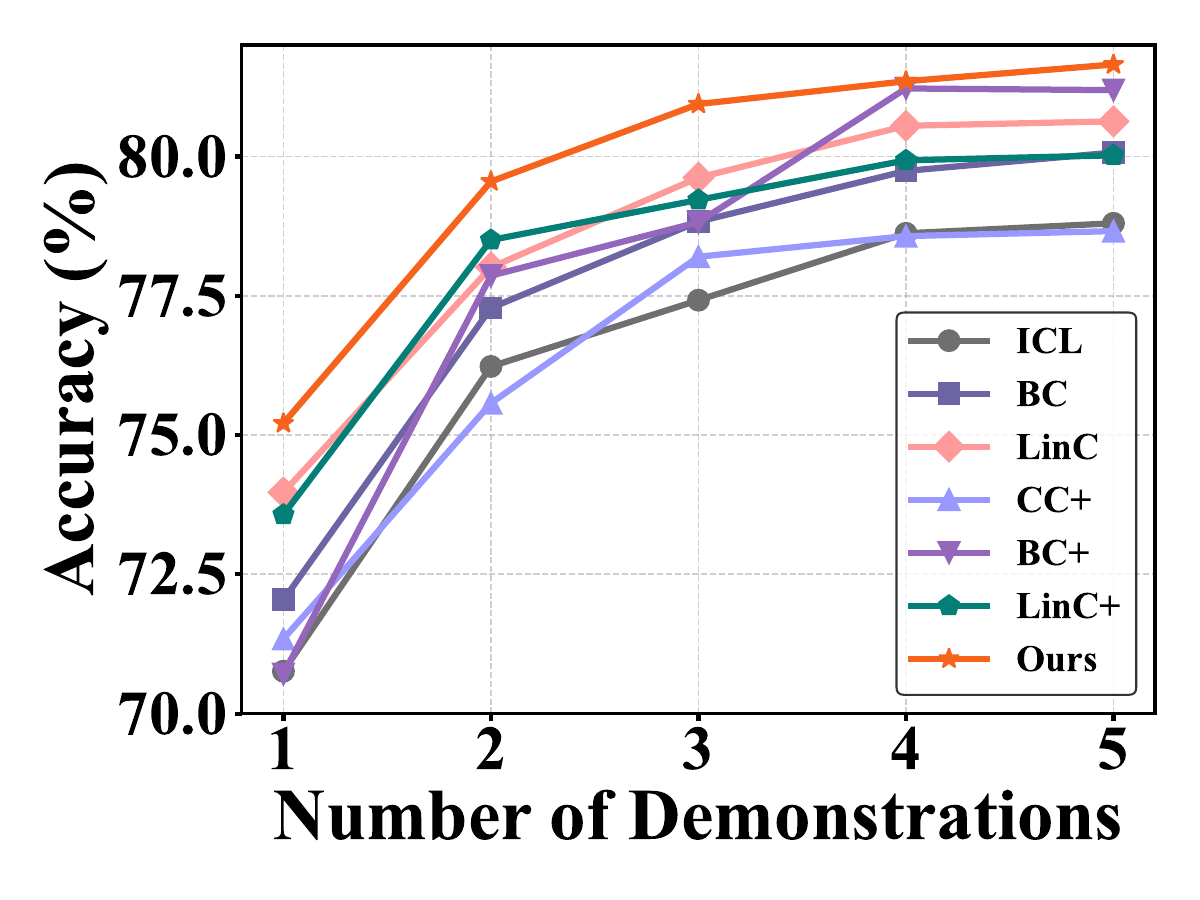}%
    \caption{Average performance across 8 datasets varies with the number of demonstrations, with other settings kept consistent as shown in Table \ref{tab:main_table}. The left panel displays results for Qwen2.5-3B, while the right panel shows results for Qwen2.5-7B.}
    \label{fig:performance-comparison}
\end{figure}
\subsection{Stability Analysis}
Previous studies \citep{xie2021explanation, rubin2022learning, liu-etal-2022-makes, lu2022fantastically, wu2023self, guo2024makes,zhang2024makes} have highlighted the sensitivity of ICL to the organization of demonstrations. In this section, we demonstrate that SC can enhance performance stability across a variety of scenarios.
\paragraph{Varying Numbers of Demonstrations}
As shown in Figure~\ref{fig:performance-comparison}, \gls{sc} again outperforms all baseline methods.  Notably, we observe that the performance of \gls{sc} under the 2-shot setting can match or even exceed that of other calibration methods in the 3-shot setting or Vanilla ICL in the 5-shot setting. Considering that the inference cost of ICL is proportional to the context length, \gls{sc} can significantly reduce computational resource consumption while maintaining comparable performance, making it particularly advantageous for real-time inference scenarios with growing demands.

\paragraph{Varying Demonstration Selecting and Ordering Strategies}
To comprehensively validate that \gls{sc} can robustly enhance the performance of \gls{icl}, we examine both demonstration selection and ordering strategies. Specifically, we evaluate three distinct demonstration selection strategies and three ordering strategies, as outlined in Section \ref{sec:Experimental Setup}. As illustrated in Figure~\ref{fig:selection}, \gls{sc} demonstrates superior performance compared to Vanilla ICL across most scenarios for both selection and ordering. Our findings indicate that selecting demonstrations more similar to the test samples generally yields better results than exclusively choosing dissimilar ones, consistent with insights reported by \citet{liu-etal-2022-makes}. Furthermore, the evaluation of ordering strategies confirms the effectiveness and reliability of \gls{sc} in enhancing model performance, as it outperforms Vanilla ICL in nearly all cases.
\begin{figure}[ht]
\includegraphics[width=0.45\textwidth]{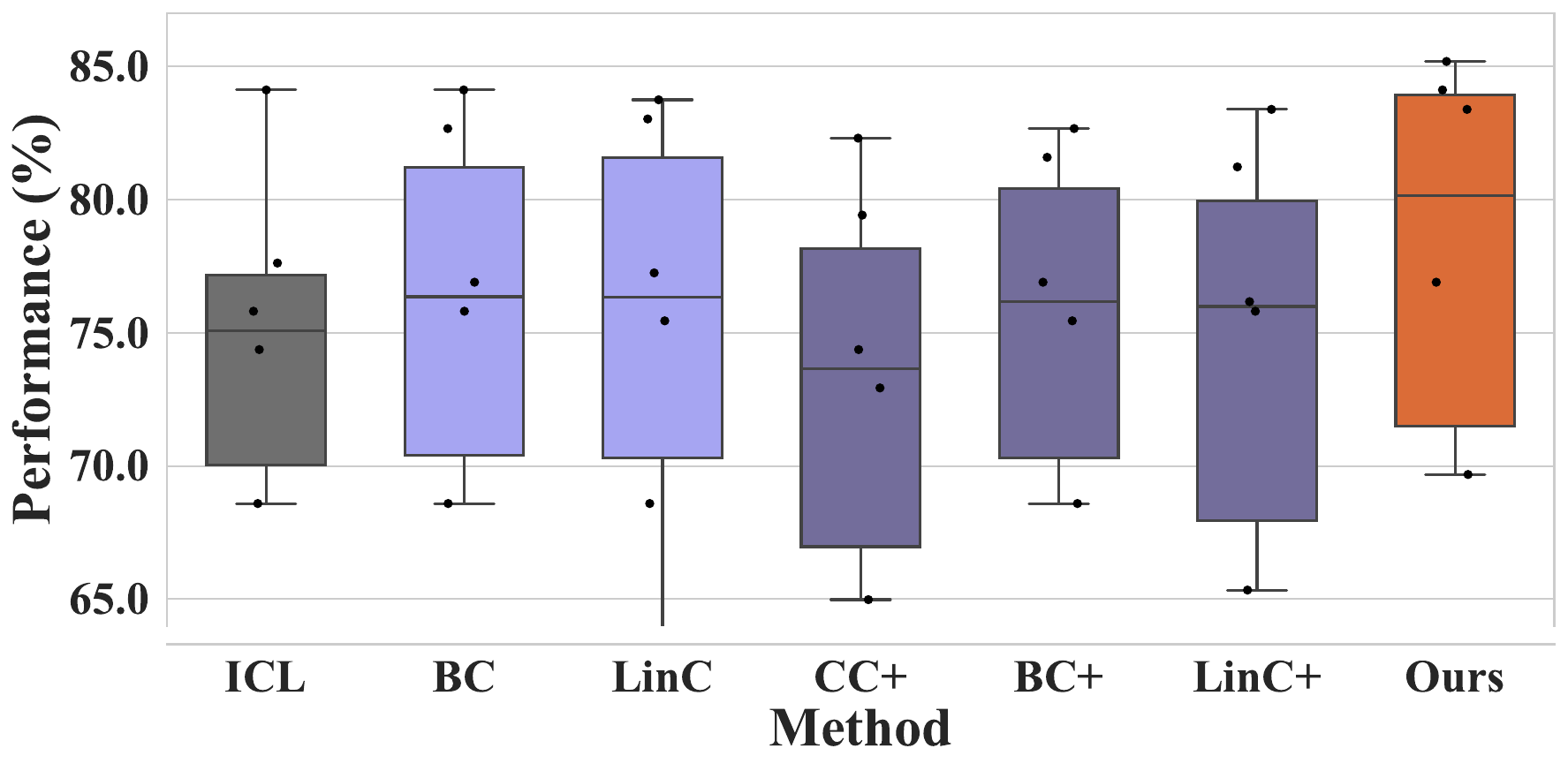}
\caption{Performance comparison of calibration methods on RTE dataset, with other settings kept consistent as shown in Table \ref{tab:main_table}. }
\label{fig:verblizer}
\end{figure}

\begin{figure}[htbp]
    \centering
    \begin{minipage}{0.45\textwidth}
        \centering
        \includegraphics[width=\linewidth]{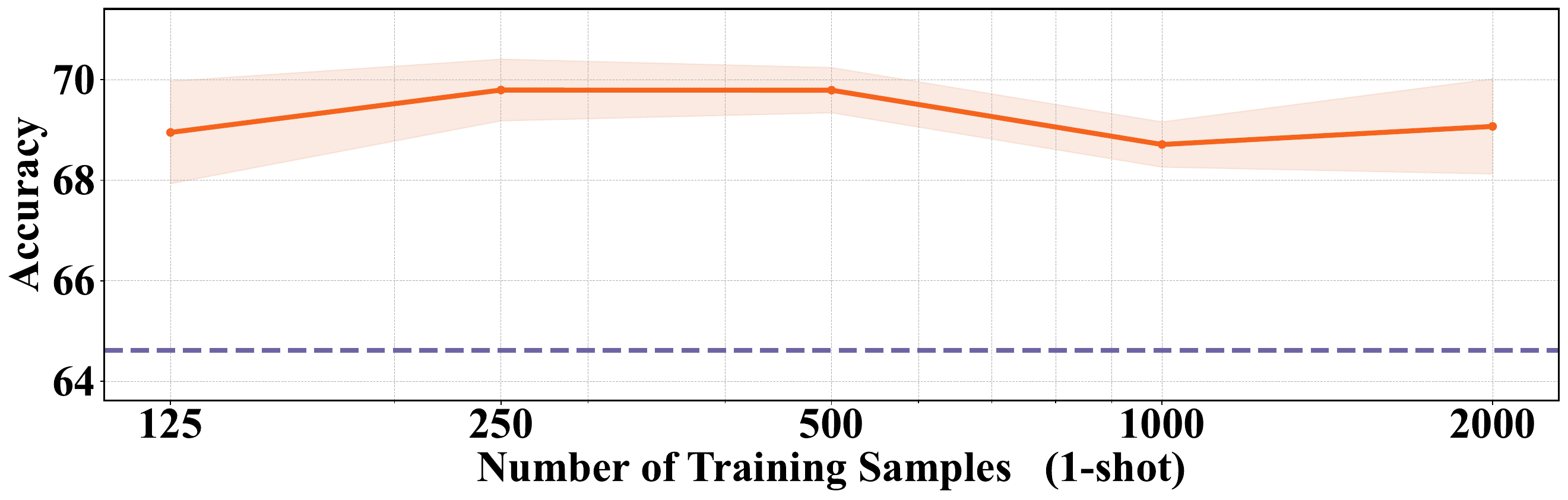}
    \end{minipage}

     \vspace{1mm} 

    \begin{minipage}{0.45\textwidth}
        \centering
        \includegraphics[width=\linewidth]{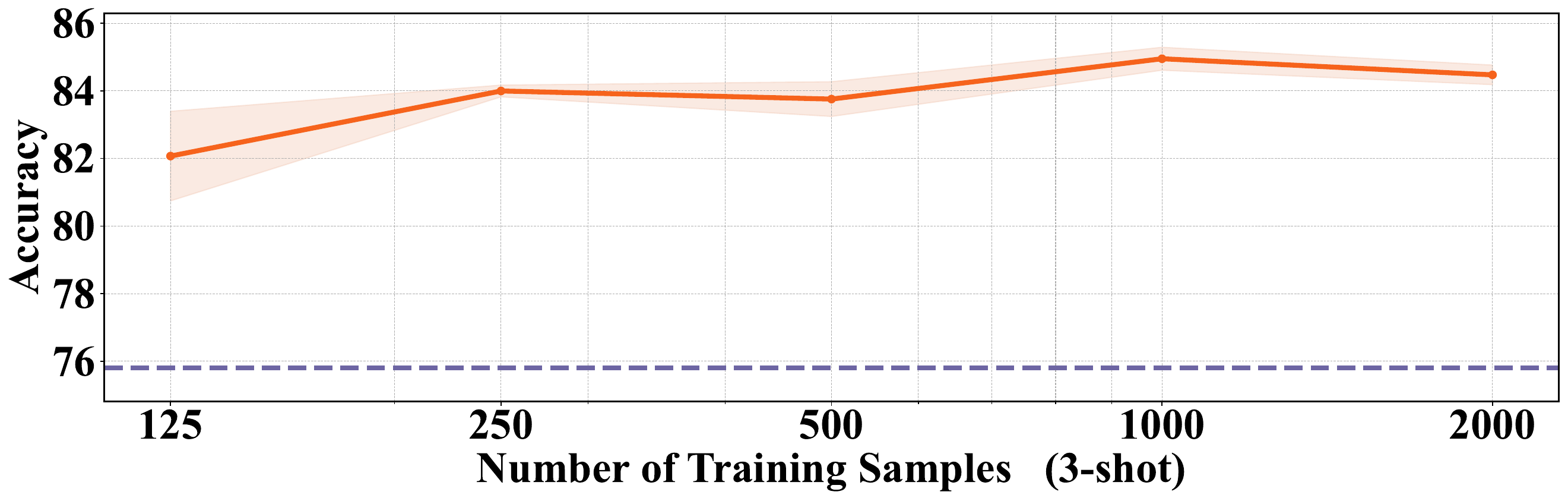}
    \end{minipage}

     \vspace{1mm} 

    \begin{minipage}{0.45\textwidth}
        \centering
        \includegraphics[width=\linewidth]{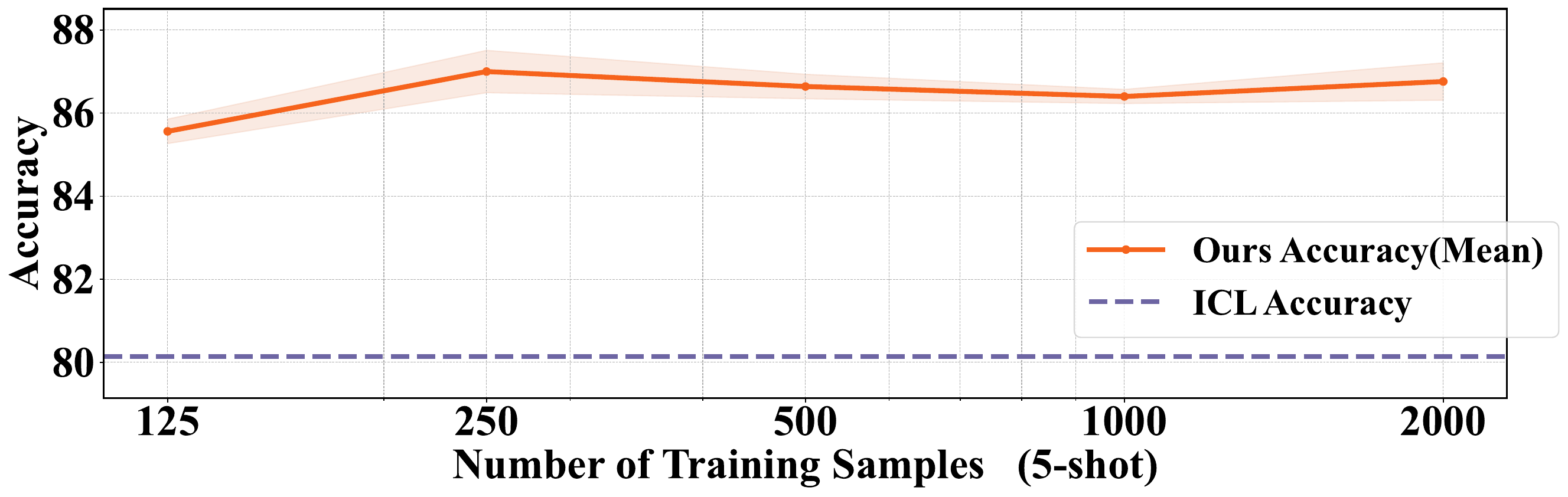}
    \end{minipage}

    \caption{Performance across varying number of training samples in 1,3 and 5-shot setting with other settings kept consistent as shown in Table \ref{tab:main_table}.}
    \label{fig:num_training_samples}
\end{figure}
\paragraph{Varying Verbalizers}
We investigate the robustness of SC to variations in verbalizer designs and find that different verbalizers influence model performance as shown in Figure~\ref{fig:verblizer}. We observed that irrespective of the verbalizer configuration, our method consistently achieves the best performance, with a median accuracy exceeding 80\%.

\paragraph{Impact of Training Set Size}
Figure~\ref{fig:num_training_samples} examines the impact of additional training samples on the performance of the RTE dataset when using Qwen2.5-3B under 1-shot, 3-shot, and 5-shot settings. The findings reveal that the model's performance remains robust across a broad range of sample sizes, consistently surpassing the ICL baseline. 
Furthermore, as the number of shots increases, the calibration model demonstrates greater stability. We hypothesize that a higher number of shots provides more diverse signals, which enhances the overall signal-to-noise ratio, thereby improving stability.

\subsection{Ablation Studies and Effectiveness Analysis}

\paragraph{Impact of Surprise Amplification}
To investigate the impact of surprise magnitude on the calibration performance of SC, we conducted ablation experiments by rescaling each dimension of the surprise vector to have an absolute value of 1, while keeping its original sign. The experimental results in Table~\ref{tab:results_split} indicate that incorporating the magnitude of surprise yields clear performance improvements for SC in most scenarios, suggesting that surprise magnitude plays a crucial role in enabling SC to accurately estimate class priors.
\begin{table}[h!]
\small
\centering
\resizebox{0.5\textwidth}{!}{
\begin{tabular}{l|cccc}
\toprule
 & SST-2 & MNLI & MRPC & QNLI \\
\midrule
w/ & $89.68 \pm 0.37$& $76.33 \pm 0.44$& $72.30 \pm 0.07$& $78.18 \pm 0.13$\\
w/o & $88.91 \pm 0.44$& $77.02 \pm 0.34$& $71.88 \pm 0.20$& $77.32 \pm 0.35$\\
\bottomrule
\end{tabular}
}
\vspace{1em} 
\resizebox{0.5\textwidth}{!}{
\begin{tabular}{l|cccc}
\toprule
 & RTE & WiC & YouTube & AIGA \\
\midrule
w/ & $84.44 \pm 0.16$& $58.05 \pm 0.34$ & $90.56 \pm 0.12$& $79.98 \pm 0.04$\\
w/o & $83.75 \pm 0.18$& $56.64 \pm 0.28$& $89.71 \pm 0.24$ & $78.00 \pm 0.18$\\
\bottomrule
\end{tabular}
}
\caption{Ablation Study. Comparison of results with (w/) and without (w/o) surprise magnitude, with other settings kept consistent as shown in Table~\ref{tab:main_table}. Results are reported as the mean ± standard deviation over three runs with three fixed random seeds.}
\label{tab:results_split}
\end{table}
\begin{figure}[t]
    \centering
        \includegraphics[width=0.25\textwidth]{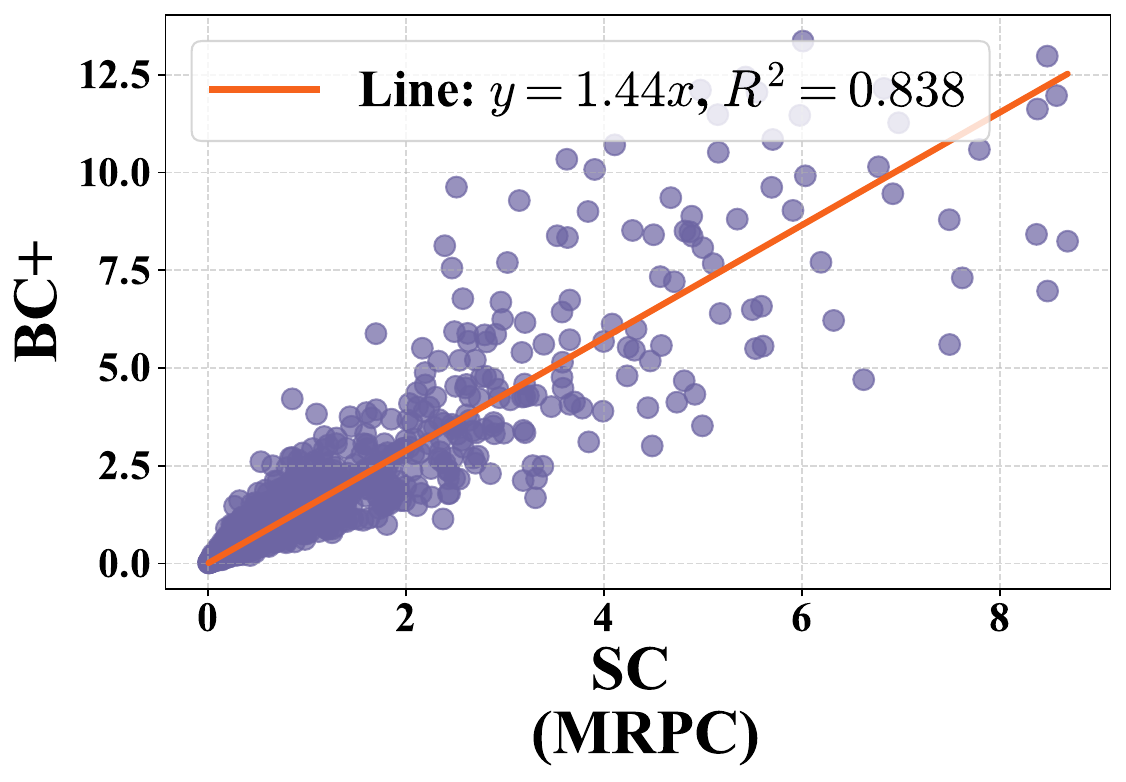}%
        \includegraphics[width=0.25\textwidth]{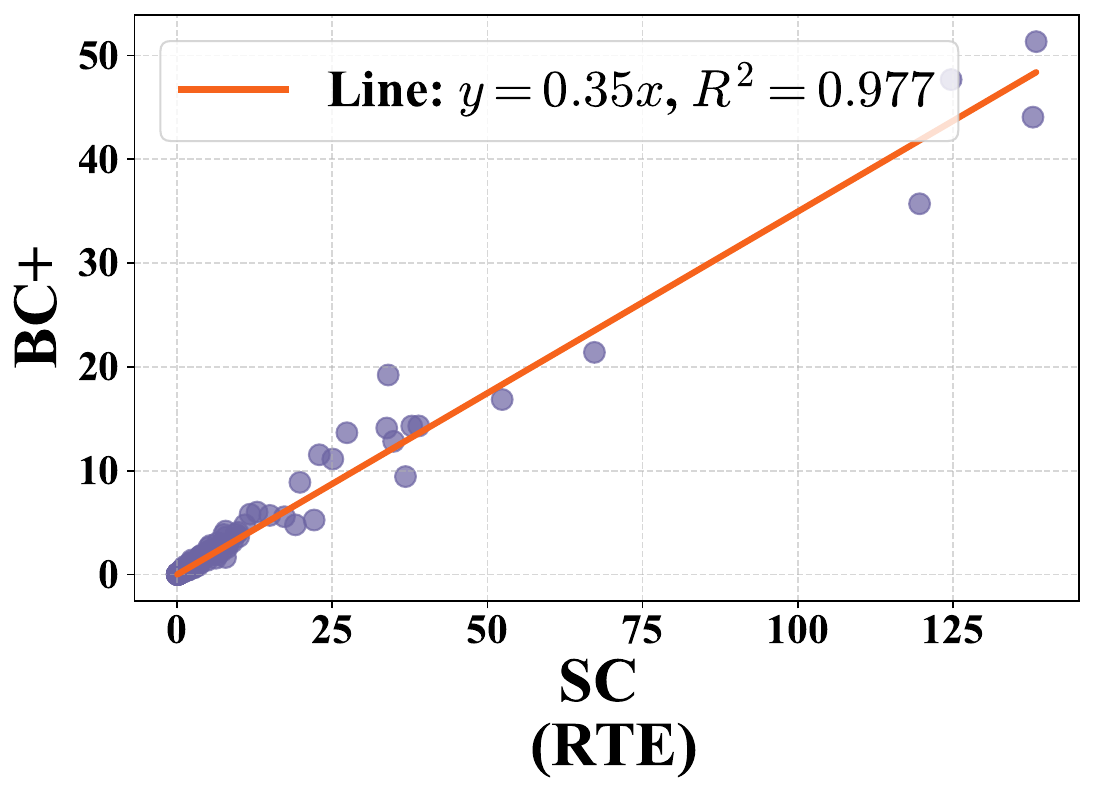}%
    \caption{Scatter plot comparing the calibrated probability ratios of SC and BC on the MRPC (left) and RTE (right) datasets. Each point represents a single evaluation instance, with other settings kept consistent as shown in Table \ref{tab:main_table}.}
    \label{fig:calib_ratio_scatter}
\end{figure}
\paragraph{Effectiveness Analysis}
To verify whether the trained SC model truly learns class priors through surprise-driven signals, we compare the calibrated probability ratio\footnote{Given the calibrated probabilities $p_\text{calib}(y=1|e)$ and $p_\text{calib}(y=0|e)$, the ratio is defined as $\frac{p_\text{calib}(y=1|e)}{p_\text{calib}(y=0|e)}$.} which directly reflects the model's relative belief in the two classes between our SC method and the Batch Calibration (BC) method.

As shown in Figure \ref{fig:calib_ratio_scatter}, there is a strong positive linear relationship between the calibrated probability ratios of SC and BC across both datasets (MRPC: $R^2 = 0.838$, RTE: $R^2 = 0.977$). This indicates that the SC model successfully captures the pattern of prior adjustment from the surprise sequence, and learns a linear decision boundary closely aligned with BC+\citep{zhou2023batch}. The linear regression slope deviates from 1, which can be attributed to the fact that both methods rely on a particular subset of data to estimate priors; differences in sample distributions can introduce a global bias, resulting in a shifted slope.
\glsresetall
\section{Conclusion}
This work adopt implicit sequential Bayesian inference as a framework for interpreting ICL, where each new demonstration is treated as an update to the prior knowledge. This interpretation allows for a more formal and principled calibration of class priors. Based on this perspective,  we introduced \gls{sc}, a novel approach to enhancing the performance and stability of \gls{icl} across diverse natural language tasks. 

Unlike existing methods, \gls{sc} does not require additional inference iterations for each input query. Instead, it uses a small number of training samples to calibrate priors, offering substantial reductions in computational cost without sacrificing performance.  We evaluated our method across eight datasets in six natural language tasks.  Our experiments demonstrate that \gls{sc}  outperforms state-of-the-art \gls{icl} bias calibration baselines. Additionally, we demonstrate that \gls{icl} effectively captures the temporal dynamics of class priors, providing more effective and robust solutions for a wide range of \gls{icl} applications.

\subsection{Limitations}
This work has several limitations. First, although our calibration performance is more stable compared to other methods, there may still be a slight decrease in performance compared to vanilla ICL in very rare cases. We speculate that this may be because implicit sequential Bayesian inference does not fully capture the behavior of ICL, rendering the surprise signal less effective in these exceptional instances. Besides, our method requires a certain amount of labeled data, making it difficult to apply in environments with extremely limited resources. Lastly, our calibration model structure is relatively simple and can be easily affected by data noise caused by factors such as model inference accuracy. 

\newpage

\bibliography{main}
\appendix

\section{Appendix}
\label{sec:appendix}

\subsection{Dataset Statistics}
\begin{table}[htbp]
\centering
\resizebox{0.5\textwidth}{!}{
\begin{tabular}{lcccccl}
\toprule
Dataset & Sentences & Classes & \multicolumn{1}{c}{|Train|} & \multicolumn{1}{c}{|Test|} & Verbizer \\
\midrule
SST-2   & 1         & 2       & 5000                      & 1821                     & negative/positive \\
MNLI    & 2         & 3       & 15000                     & 9815                     & no/maybe/yes      \\
MRPC    & 2         & 2       & 4075                      & 1724                     & no/yes            \\
QNLI    & 2         & 2       & 5000                      & 5463                     & no/yes            \\
RTE     & 2         & 2       & 2490                      & 277                      & no/yes            \\
WiC     & 3         & 2       & 5428                      & 1400                     & false/true        \\
YouTube & 1         & 2       & 1564                      & 392                      & truthful/deceptive \\
AIGA    & 1         & 2       & 4320                      & 5732                     & 0/1               \\
\bottomrule
\end{tabular}
}
\caption{Details of the dataset used for evaluation,"Sentences" denotes the number of segments in the input, while "Classes" refers to the number of categories in the label space, "|Train|" and "|Test|" denotes the number of test samples. When the labels of the test set are not publicly available, we use the validation set as the test set. For all datasets, we use a verbalizer that is semantically aligned with its label space.}
\label{tab:dataset_info}
\end{table}
\subsection{Implementation Details}
\begin{itemize}
    \item \textbf{BC} (Batch Calibration) \citep{zhou2023batch}: We reproduce the BC baseline by using the batch of all testing samples to estimate the class prior.
    \item \textbf{LinC} (Linear Probe Calibration) \citep{abbas2024enhancing}: We follow the original implementation of LinC and using the same training sample size as \gls{sc} method.
    \item \textbf{CC+} (Contextual Calibration) \citep{zhao2021calibrate}: We adhere to the original implementation of CC and compute the mean of the log-probabilities over three content-free tokens—‘N/A’, ‘’, and ‘[MASK]’—as the test sample within the predefined template.
    \item \textbf{BC+} We extend the BC baseline by using a fixed batch of labeled samples to estimate the class prior for each new input, selecting five samples for each class and combining them into a single batch.
    \item \textbf{LinC+} We extend the LinC baseline by using a fixed number of labeled samples to estimate the class prior for each new input, selecting five samples for each class and combining them into a training set.
    \item  \textbf{\gls{sc}} In our preliminary experiments, we evaluated several common backbone architectures, including GRU, LSTM, vanilla RNN, and Transformer Encoder. The results are summarized in the following table(Qwen2.5-3B;RTE;3-shot):
    \begin{table}[ht]
    \centering
    \begin{tabular}{l c}
    \toprule
    \textbf{Backbone} & \textbf{Accuracy} \\
    \midrule
    GRU         & 84.47\% \\
    LSTM        & 83.95\% \\
    RNN         & 84.47\% \\
    Transformer & 84.47\% \\
    \bottomrule
    \end{tabular}
    \caption{Accuracy comparison of different backbones.}
    \label{tab:backbone_accuracy}
    \end{table}
    As shown above, all backbones achieved very similar accuracy, indicating that the choice of backbone had minimal influence on the final calibration performance. We selected GRU due to its simpler structure and ease of training. The BSC model was trained for 200 epochs using the Adam optimizer with a learning rate of 1e-4. Hyperparameters were selected based on early-stage experiments on the MRPC dataset, using 20\% of its training data as a validation set. These selected hyperparameters were then applied unchanged to all other datasets. Given the relatively simple nature of the model and datasets, we found this level of tuning to be sufficient.

    \item \textbf{Model Predicted Probabilities}: In the general practice of ICL, due to the influence of the in-context demonstrations, when the model predicts the label for a query, the sum of probabilities for all tokens in the label space tends to be very close to 1, while the probabilities of other tokens in the vocabulary are close to 0. Therefore, to understand the model’s final prediction preference, we can focus solely on the limited tokens within the label space. Consequently, we only decode the hidden states of each delimiter into the tokens of the label space. (Only keep the logits of the label tokens.) So far, we have assumed that each label string is encoded as a single token. However, our approach can also be applied if some or all labels are encoded as multiple tokens. In essence, we continue to measure only the probability the model assigns to the first token of each label, making the (fairly harmless) assumption that the first (or only) token that each label is encoded to is unique among labels. We believe this is justified, as, given the first token for a label, the model should near-deterministically predict the remaining tokens, i.e. all the predictive information is contained in the first token the model predicts for a label. For example, for the YouTube dataset, the label ‘truthful’ is encoded by the Qwen tokenizer as two tokens, [truth, ful]. We only use the probability assigned to [truth] to assign probabilities to ‘truthful’, and ignore any predictions for [ful].
\end{itemize}

\subsection{Additional Experiments}
\begin{figure*}[htb]
    \centering
    \resizebox{\textwidth}{!}{
        \includegraphics[width=0.25\textwidth]{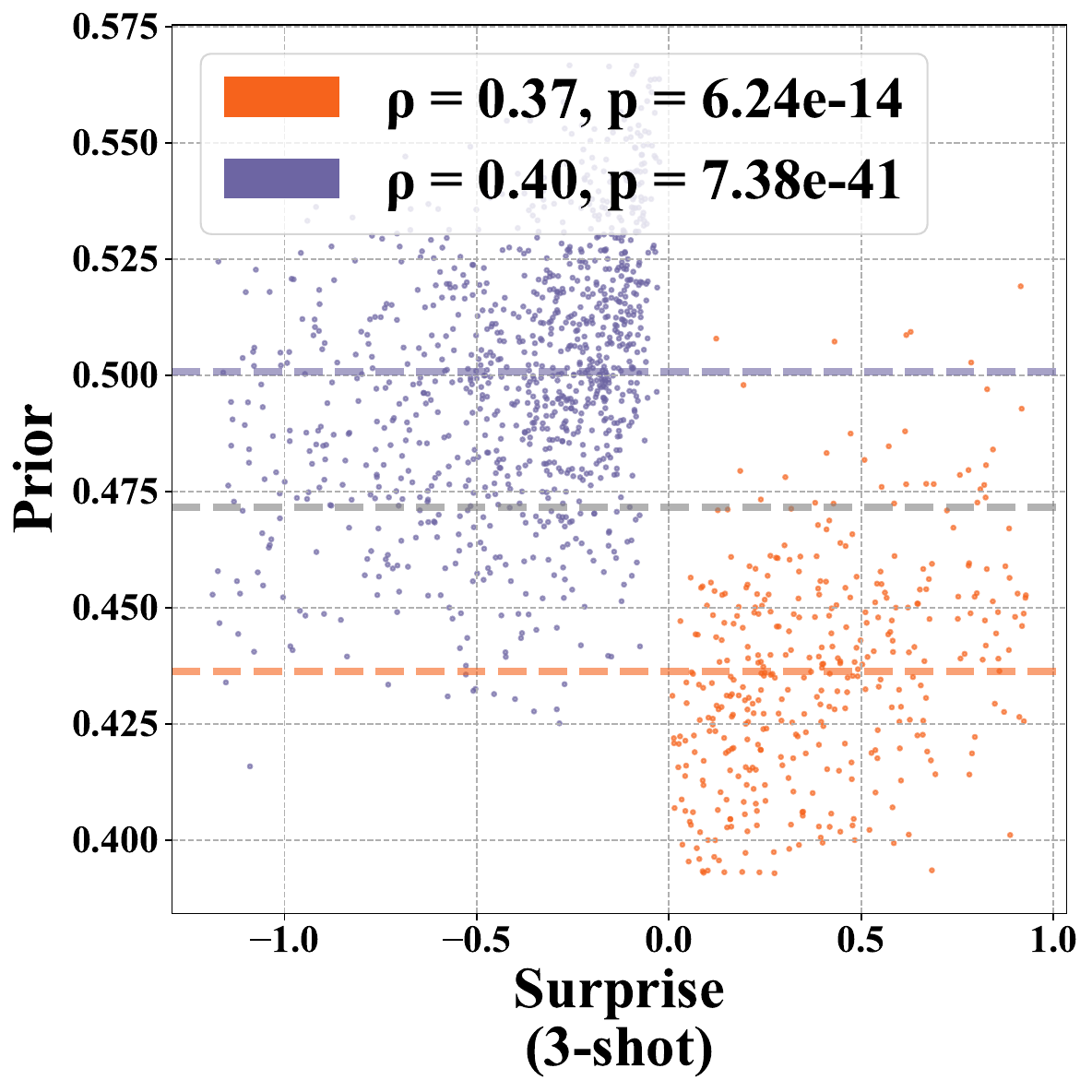}%
        \includegraphics[width=0.25\textwidth]{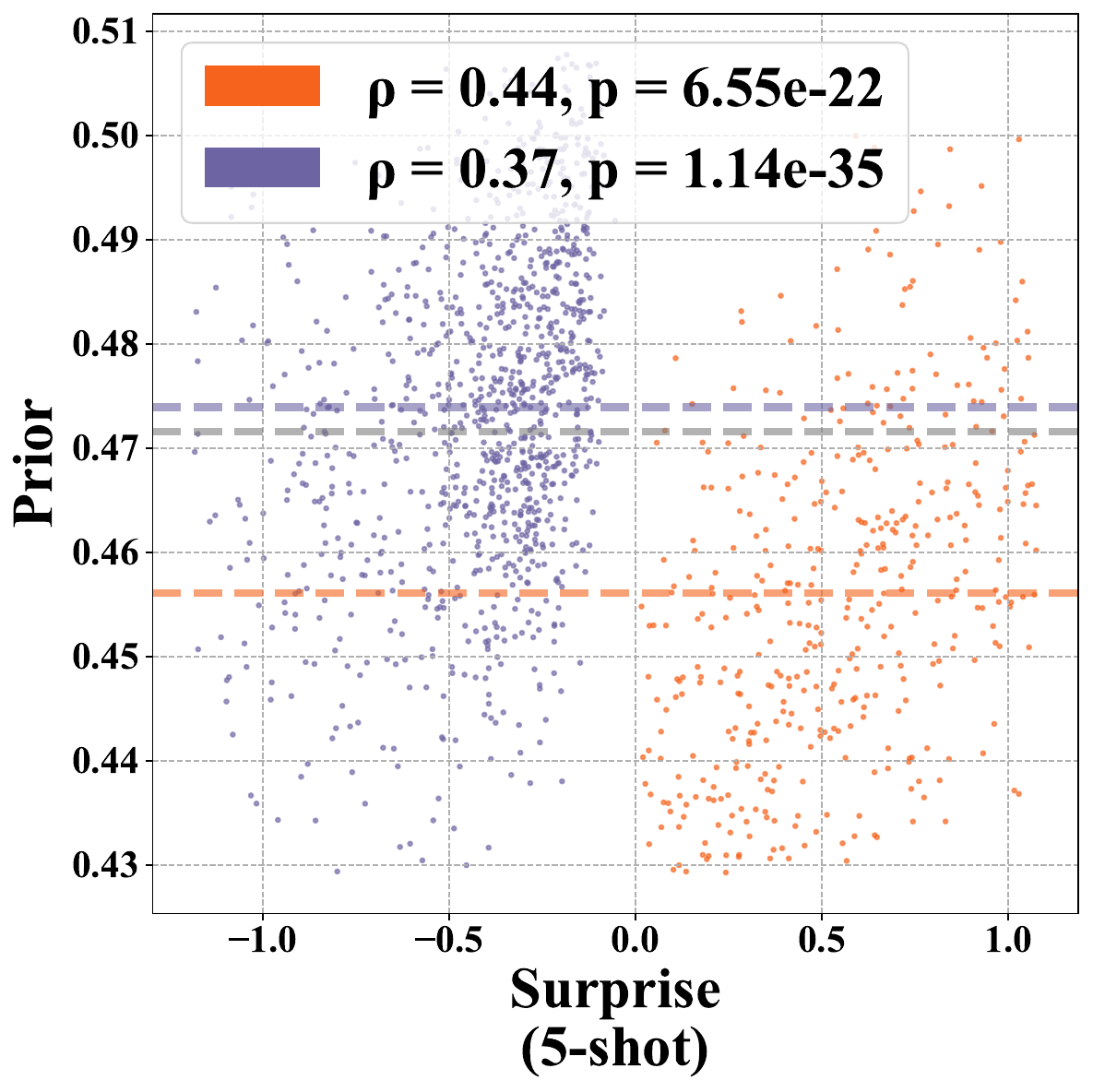}%
        \includegraphics[width=0.25\textwidth]{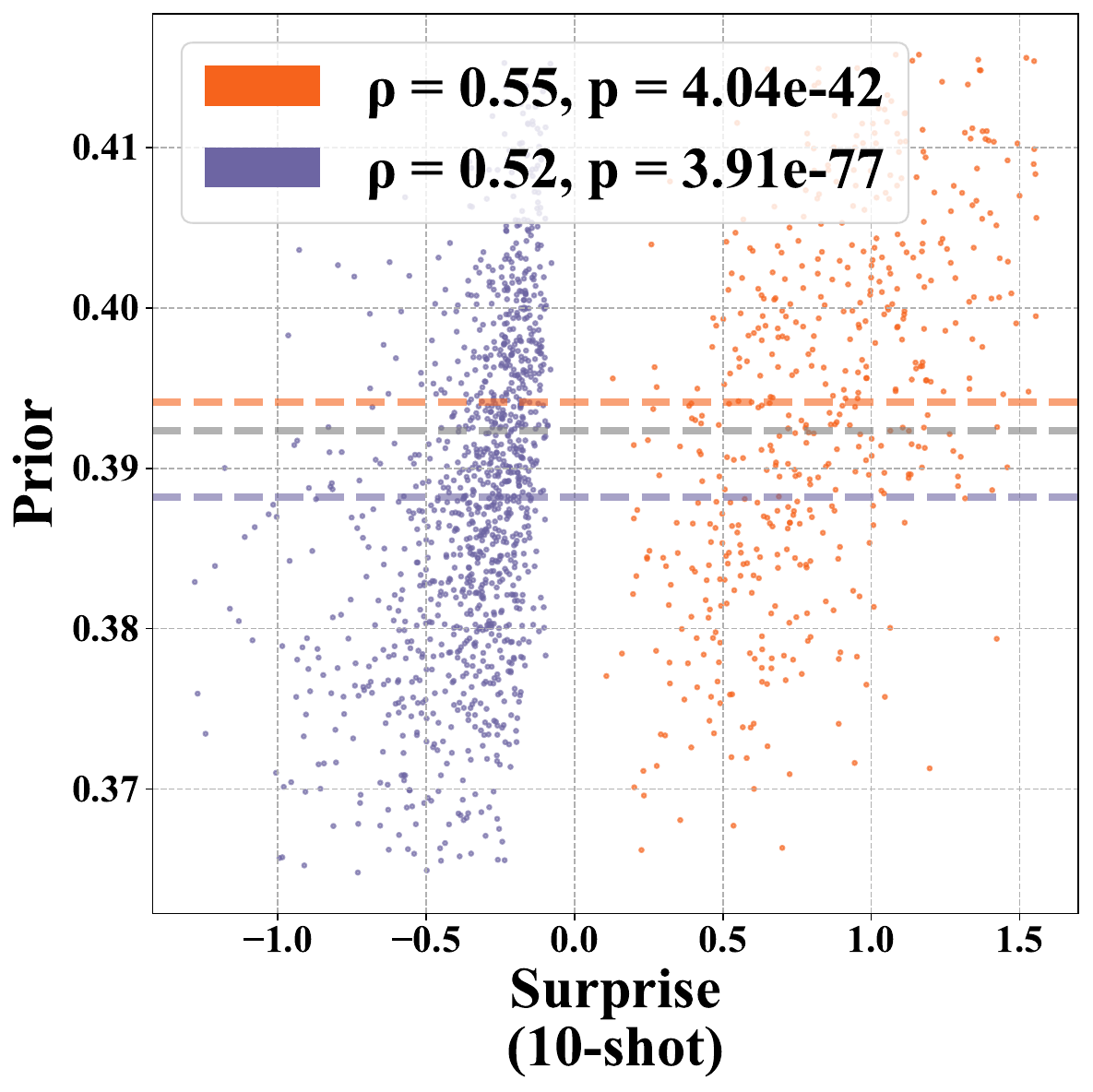}%
        \includegraphics[width=0.25\textwidth]{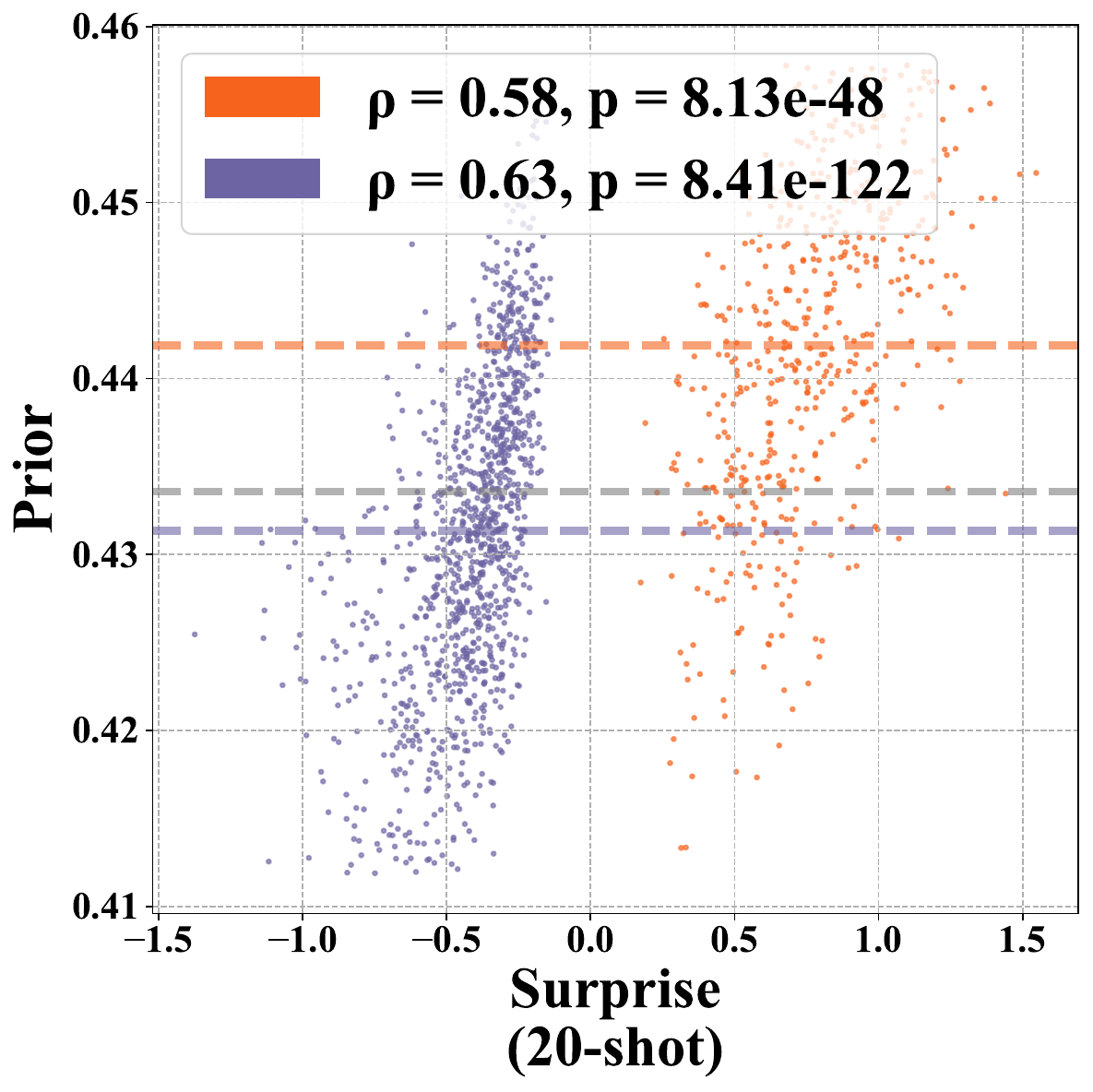}%
    }
    \caption{Spearman correlation between surprise ($-\log p(y|e, D)$) and the prior probability of the positive class across various in-context settings (3-, 5-, 10-, 20-shot) with MRPC dataset. Each scatter plot shows the relationship for positive (orange) and negative (purple) demonstration insertions, with corresponding Spearman $\rho$ and $p$-values.
    The gray dashed line marks the estimated prior before insertion, while orange and purple dashed lines indicate the average prior after inserting positive and negative demonstration, respectively. All priors are estimated by repeated sampling using the BC method (described in Section 4). Results show the effect of anti-recency bias diminishes with increasing numbers of in-context demonstrations, indicating it is primarily small-sample phenomenon.}
    \label{fig:anti-recency bias}
\end{figure*}

\begin{figure*}[ht]
\centering
\begin{subfigure}[b]{0.8\textwidth}
    \centering
    \includegraphics[width=\textwidth]{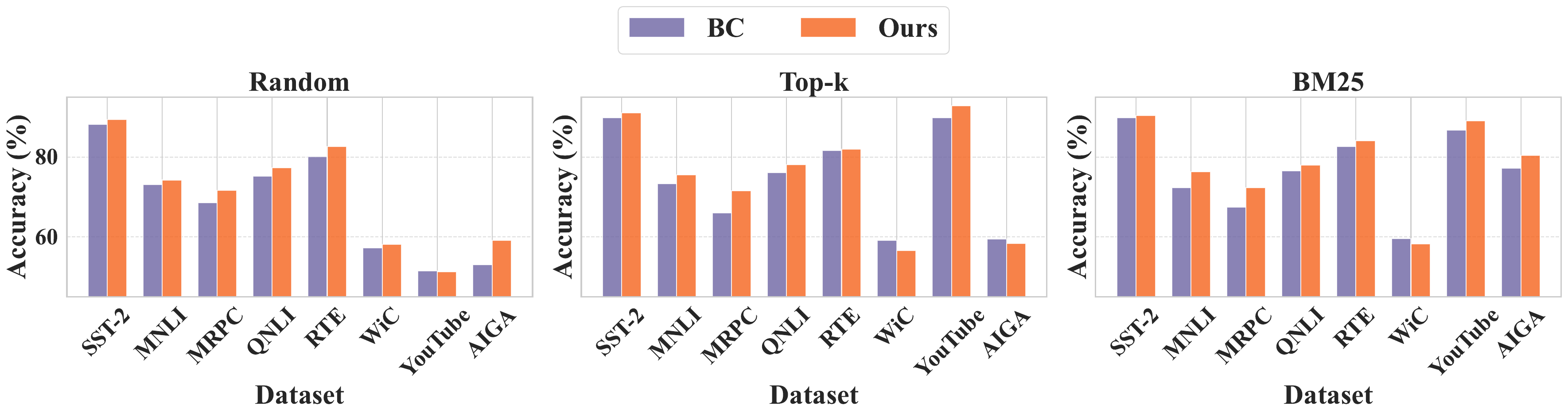}
    \caption{Comparison of \gls{sc} and BC.}
    \label{fig:selection_bc}
\end{subfigure}

\begin{subfigure}[b]{0.8\textwidth}
    \centering
    \includegraphics[width=\textwidth]{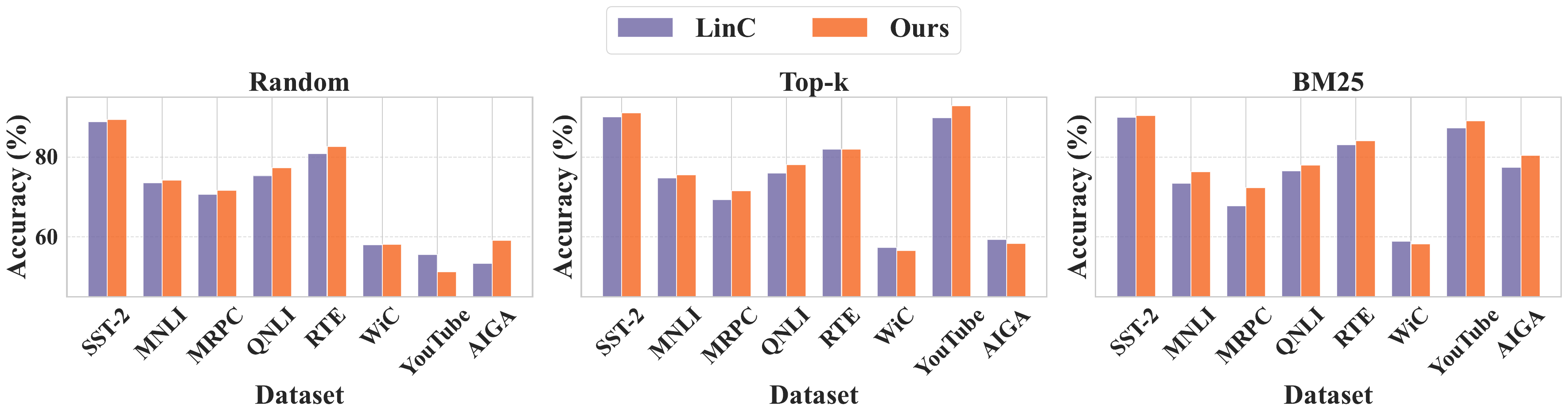}
    \caption{Comparison of \gls{sc} and LinC.}
    \label{fig:selection_linc}
\end{subfigure}

\begin{subfigure}[b]{0.8\textwidth}
    \centering
    \includegraphics[width=\textwidth]{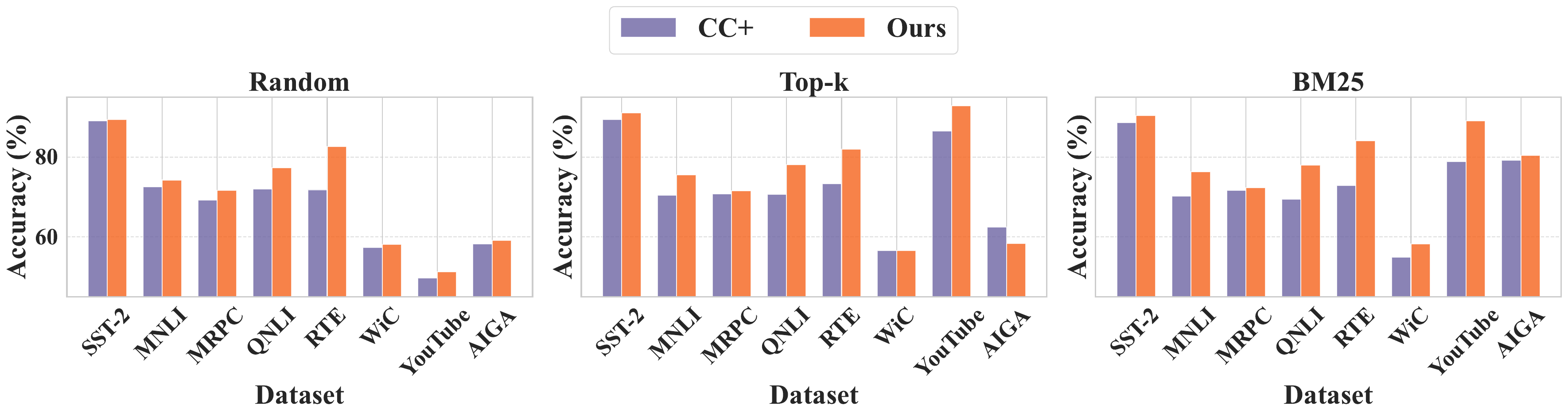}
    \caption{Comparison of \gls{sc} and CC+.}
    \label{fig:selection_ccplus}
\end{subfigure}

\begin{subfigure}[b]{0.8\textwidth}
    \centering
    \includegraphics[width=\textwidth]{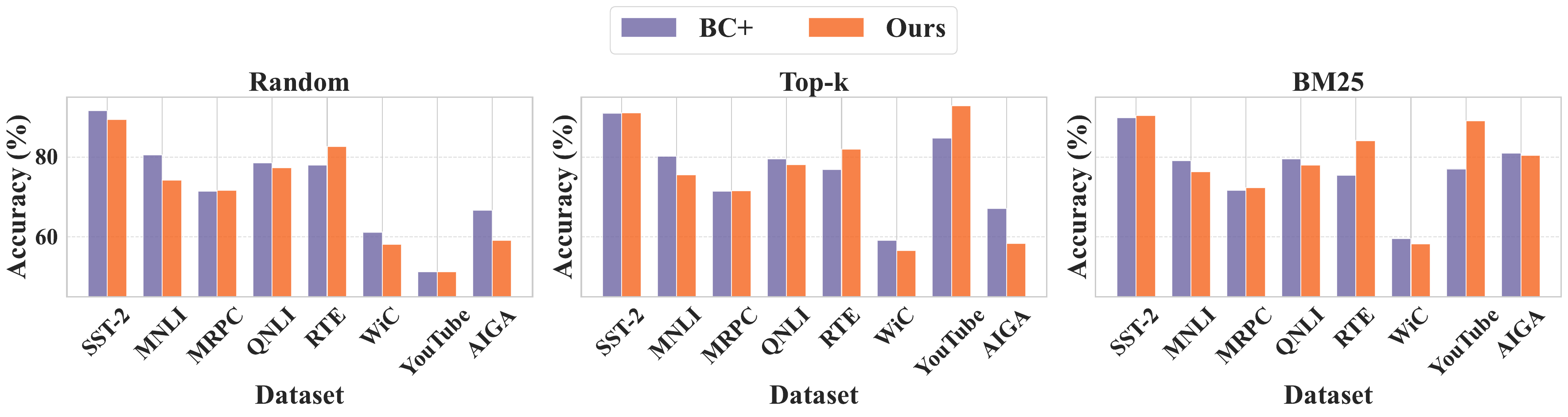}
    \caption{Comparison of \gls{sc} and BC+.}
    \label{fig:selection_bcplus}
\end{subfigure}

\begin{subfigure}[b]{0.8\textwidth}
    \centering
    \includegraphics[width=\textwidth]{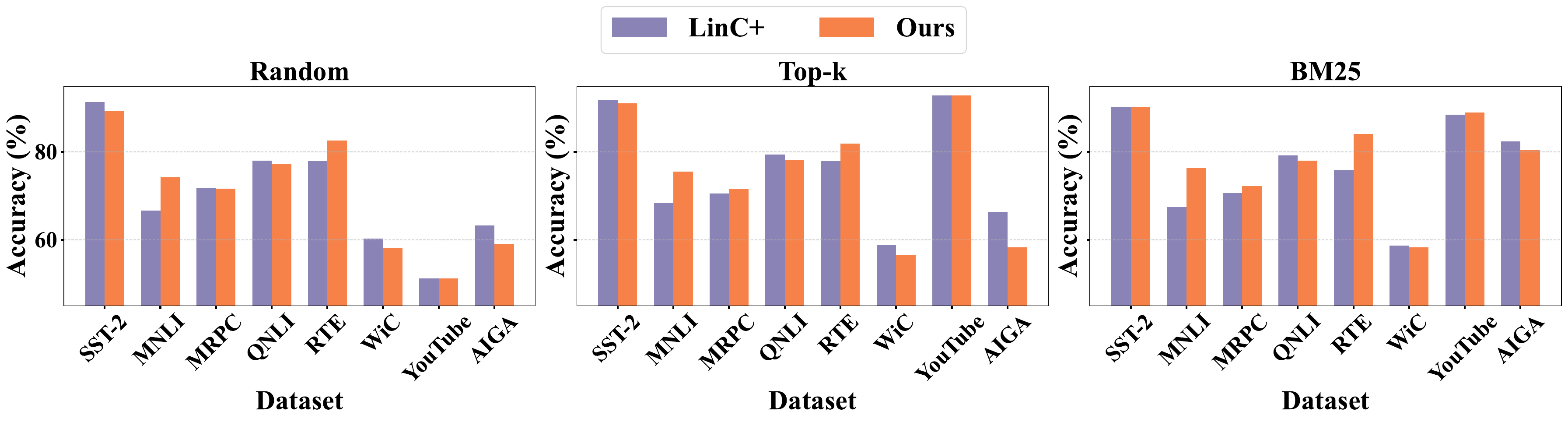}
    \caption{Comparison of \gls{sc} and LinC+.}
    \label{fig:selection_lincplus}
\end{subfigure}

\caption{Accuracy comparisons between \gls{sc} and various methods (BC, LinC, CC+, BC+, LinC+) across three demonstration selection strategies. Other settings are consistent with those shown in Table \ref{tab:main_table}.}
\label{fig:selection_comparison}
\end{figure*}

\begin{figure*}[ht]
\centering
\begin{subfigure}[b]{0.8\textwidth}
    \centering
    \includegraphics[width=\textwidth]{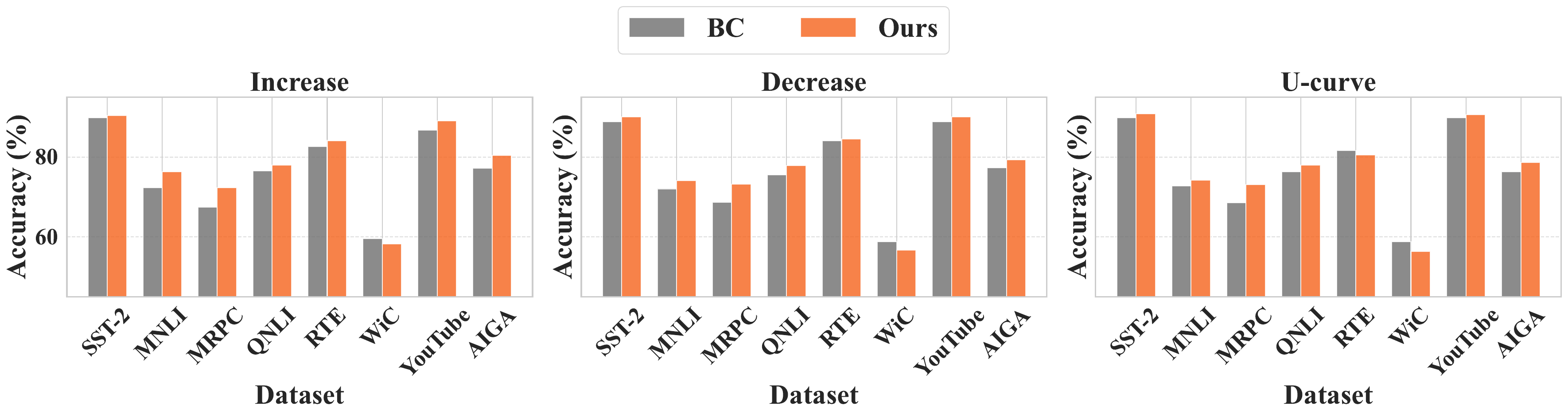}
    \caption{Comparison of \gls{sc} and BC.}
    \label{fig:order_bc}
\end{subfigure}

\begin{subfigure}[b]{0.8\textwidth}
    \centering
    \includegraphics[width=\textwidth]{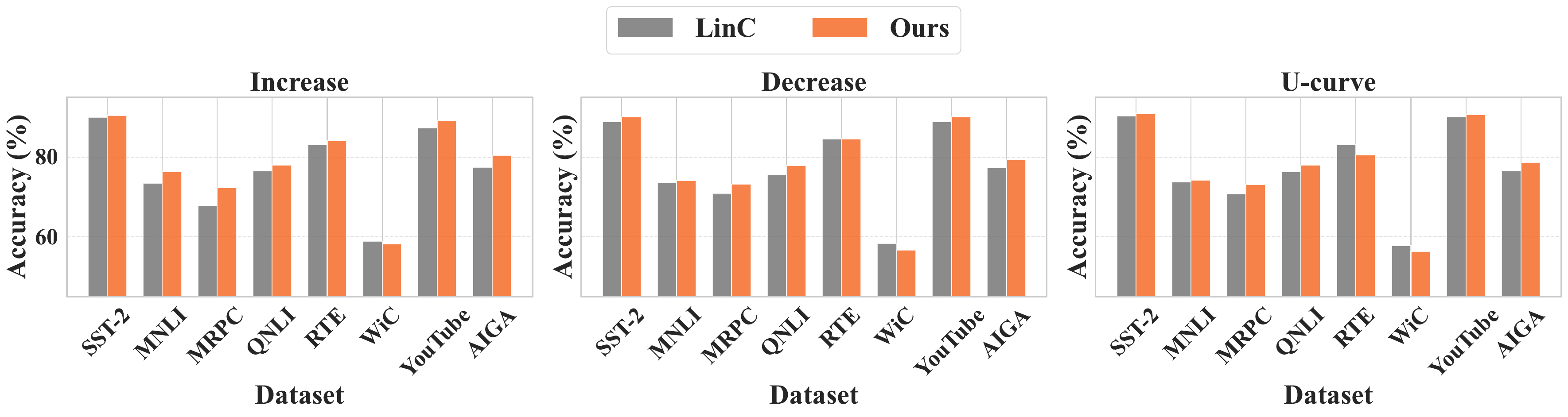}
    \caption{Comparison of \gls{sc} and LinC.}
    \label{fig:order_linc}
\end{subfigure}

\begin{subfigure}[b]{0.8\textwidth}
    \centering
    \includegraphics[width=\textwidth]{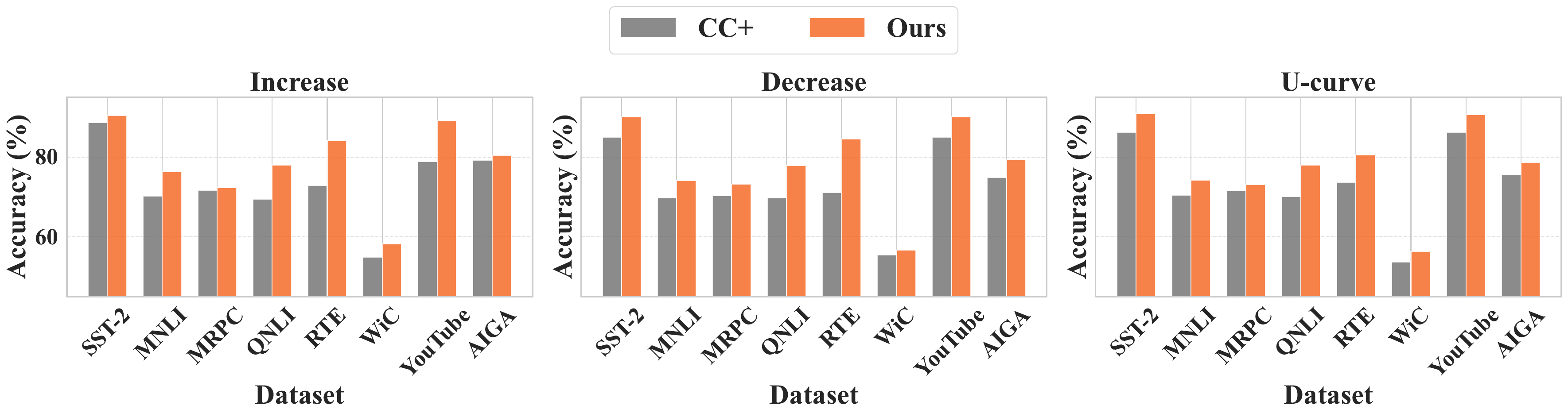}
    \caption{Comparison of \gls{sc} and CC+.}
    \label{fig:order_ccplus}
\end{subfigure}

\begin{subfigure}[b]{0.8\textwidth}
    \centering
    \includegraphics[width=\textwidth]{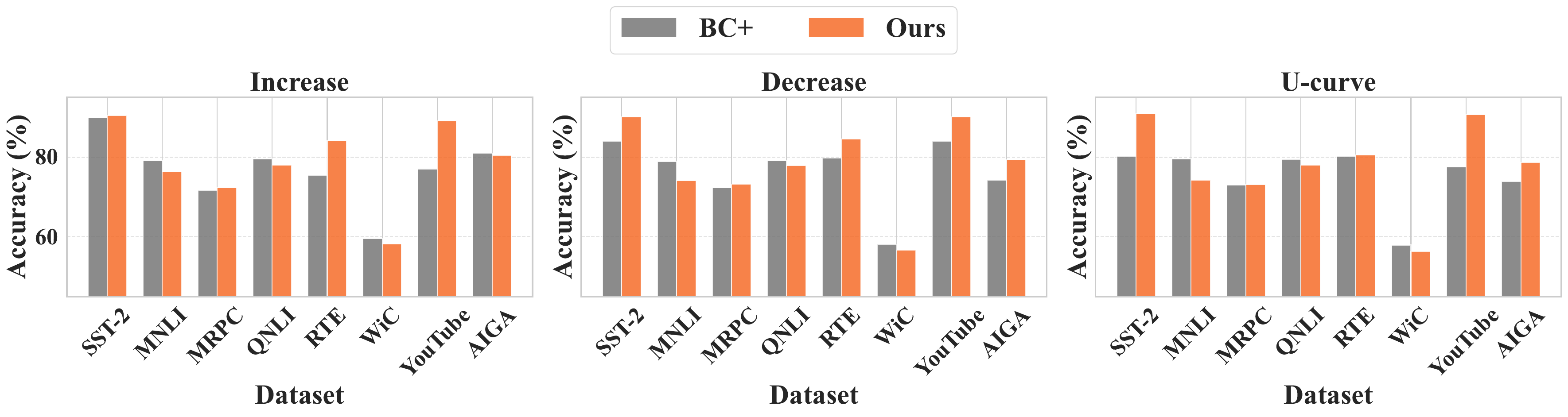}
    \caption{Comparison of \gls{sc} and BC+.}
    \label{fig:order_bcplus}
\end{subfigure}

\begin{subfigure}[b]{0.8\textwidth}
    \centering
    \includegraphics[width=\textwidth]{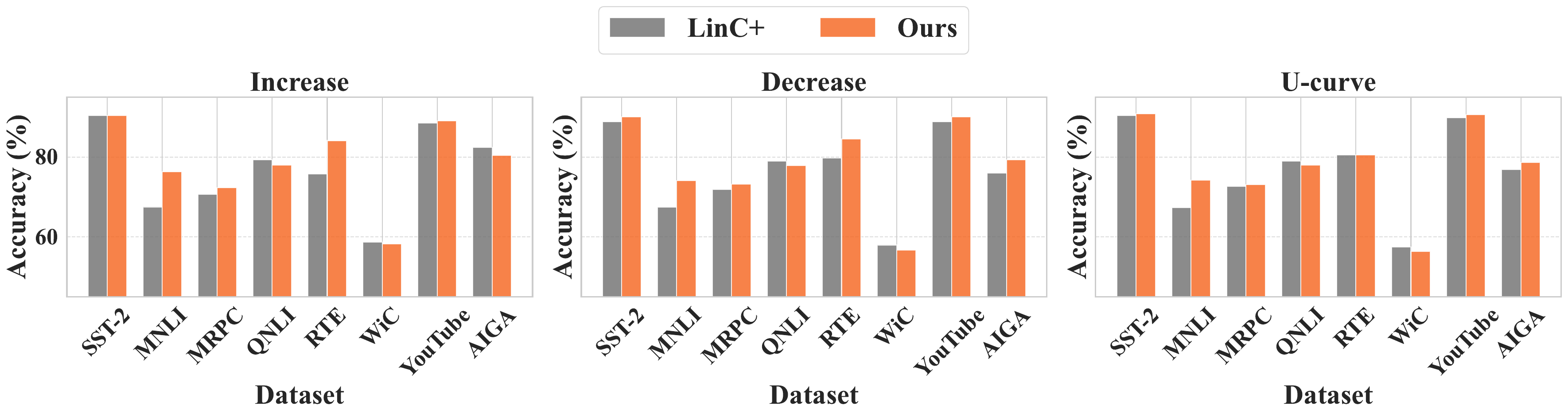}
    \caption{Comparison of \gls{sc} and LinC+.}
    \label{fig:order_lincplus}
\end{subfigure}

\caption{Accuracy comparisons between \gls{sc} and various methods (BC, LinC, CC+, BC+, LinC+) across three demonstration ordering strategies. Other settings remain consistent with those specified in Table \ref{tab:main_table}.}
\label{fig:comparison}
\end{figure*}

Figure~\ref{fig:anti-recency bias} illustrates that the effect of anti-recency bias diminishes as the number of demonstrations.
Table~\ref{tab:many-shot result} illustrates the experimental results on more shot.
Figure~\ref{fig:performance-comparison} illustrates the experimental results, providing evidence for the validity of the theorem introduced in Section~\ref{sec:seq-updates}.  
Table~\ref{tab:main} summarizes the experimental outcomes under 1-shot, 2-shot, 4-shot, and 5-shot settings.  
Figures~\ref{fig:selection_comparison} and \ref{fig:comparison} present a comparative analysis between the \gls{sc} approach and the baseline method.

\begin{table*}[ht]
\scriptsize
\centering
\begin{minipage}{0.5\textwidth}
\centering
\begin{tabularx}{\linewidth}{lccccccc}
\toprule
\multicolumn{8}{c}{\textbf{10-shot Results}} \\
\textbf{Dataset} & \textbf{ICL} & \textbf{BC} & \textbf{LinC} & \textbf{CC+} & \textbf{BC+} & \textbf{LinC+} & \textbf{Ours} \\
\midrule
MRPC & 73.62 & 72.11 & 72.93 & 71.01 & 74.02 & 73.97 & \textbf{73.86} \\
WiC  & 56.14 & 58.21 & 57.07 & 57.14 & \textbf{58.79} & 58.64 & 57.86 \\
MNLI & 71.19 & 77.83 & \textbf{78.07} & 74.77 & 77.26 & 73.29 & 77.91 \\
QNLI & 77.74 & 78.32 & 78.23 & 66.57 & 78.16 & \textbf{78.80} & 78.73 \\
\midrule
\textbf{Avg.} & 69.67 & 71.62 & 71.58 & 67.37 & \textbf{72.06} & 71.18 & \textbf{72.09} \\
\bottomrule
\end{tabularx}
\end{minipage}\hfill
\begin{minipage}{0.5\textwidth}
\centering
\begin{tabularx}{\linewidth}{lccccccc}
\toprule
\multicolumn{8}{c}{\textbf{15-shot Results}} \\
\textbf{} & \textbf{ICL} & \textbf{BC} & \textbf{LinC} & \textbf{CC+} & \textbf{BC+} & \textbf{LinC+} & \textbf{Ours} \\
\midrule
 & 74.38 & 72.05 & 73.91 & 70.66 & 74.14 & 74.37 & \textbf{74.43} \\
  & 57.28 & 58.64 & 57.35 & 56.85 & 58.50 & 58.07 & \textbf{58.14} \\
 & 70.80 & 77.25 & 77.91 & 74.31 & \textbf{78.68} & 72.59 & 78.23 \\
 & 76.64 & 77.41 & 77.41 & 63.21 & 75.98 & 76.45 & \textbf{78.01} \\
\midrule
\textbf{} & 69.78 & 71.34 & 71.65 & 66.26 & 71.83 & 70.37 & \textbf{72.20} \\
\bottomrule
\end{tabularx}
\end{minipage}

\caption{Comparison of 10-shot and 15-shot Results on different datasets. 
We have conducted additional experiments on the most challenging datasets (MRPC, WiC, MNLI, QNLI). These results confirm that Surprise Calibration (BSC) continues to outperform or match existing baselines under increased shot settings, maintaining robustness and effectiveness across harder tasks. Importantly, we also observed that performance improvements begin to saturate as the number of demonstrations increases beyond 5. This phenomenon is consistent with findings reported by \citet{agarwal2024many}. For example, the improvement from 10-shot to 15-shot is often marginal (e.g., 72.09 → 72.20 in average accuracy). This trend suggests that SC is already effective in leveraging limited demonstration context, and adding more demonstrations yields only minor additional benefit, likely due to information redundancy in processing extended contexts.}
\label{tab:many-shot result}
\end{table*}

\begin{table*}[ht]
\centering
\begin{subtable}{0.45\textwidth}
    \centering
    \resizebox{\textwidth}{!}{
\begin{tabular}{@{\extracolsep{\fill}} l c c c c c c c c}
\toprule
\multirow{2}{*}{\textbf{DataSet}} & \multirow{2}{*}{\textbf{LM}} & \multicolumn{7}{c}{\textbf{Method}} \\
 & &  \textbf{ICL} & \textbf{BC} & \textbf{LinC} & \textbf{CC+} & \textbf{BC+} & \textbf{LinC+} & \textbf{Ours} \\
\midrule
SST-2 & Qwen2.5-3B & 78.52\%          & 80.01\%          & 78.97\%          & 71.66\%     & \textbf{81.16\%} & 80.01\%          & 76.57\%\\
& Qwen2.5-7B & 85.23\%          & 85.61\%          & 85.72\%          & 88.19\%     & 88.25\%          & 86.60\%          & \textbf{91.32\%}\\
\midrule
MNLI & Qwen2.5-3B & 50.03\%          & 54.33\%          & 57.99\%          & 58.91\%     & \textbf{70.01\%} & 54.87\%          & 68.04\%\\
& Qwen2.5-7B & 59.48\%          & 62.73\%          & 68.48\%          & 70.92\%     & \textbf{77.78\%} & 60.33\%          & 75.86\%\\
\midrule
MRPC & Qwen2.5-3B & 67.42\%          & 63.83\%          & 68.23\%          & 67.25\%     & 70.78\%          & 70.67\%          & \textbf{72.34\%}\\
& Qwen2.5-7B & 60.99\%          & 59.59\%          & 68.52\%          & 69.27\%     & 69.74\%          & 69.79\%          & \textbf{71.19\%}\\
\midrule
QNLI & Qwen2.5-3B & 63.82\%          & 70.24\%          & 70.25\%          & 56.94\%     & \textbf{74.42\%} & 74.37\%          & 71.84\%\\
& Qwen2.5-7B & 72.49\%          & 76.64\%          & 76.71\%          & 60.64\%     & \textbf{77.21\%} & 76.73\%          & 76.64\%\\
\midrule
RTE & Qwen2.5-3B & 64.62\%          & 67.51\%          & 67.51\%          & 62.81\%     & 72.20\%          & \textbf{70.03\%} & 70.04\%\\
& Qwen2.5-7B & 69.31\%          & 71.84\%          & 72.20\%          & 71.84\%     & 74.37\%          & 74.73\%          & \textbf{75.45\%}\\
\midrule
WiC & Qwen2.5-3B & 51.00\%          & \textbf{55.50\%} & 54.64\%          & 50.29\%     & 53.29\%          & 53.57\%          & 55.35\%\\
& Qwen2.5-7B & 56.00\%          & 57.35\%          & 57.50\%          & 54.86\%     & 58.21\%          & \textbf{58.50\%} & 56.07\%\\
\midrule
YouTube & Qwen2.5-3B & 89.54\%          & \textbf{89.80\%} & 89.54\%          & 84.94\%     & 62.50\%          & 89.29\%          & 88.80\%\\
& Qwen2.5-7B & \textbf{89.79\%} & \textbf{89.79\%} & \textbf{89.79\%} & 84.95\%     & 65.56\%          & 89.79\%          & 89.54\%          
\\
\midrule
AIGA & Qwen2.5-3B & \textbf{72.82\%} & 72.82\%          & 72.82\%          & 73.01\%     & 56.43\%          & 71.35\%          & 72.47\%          
\\
& Qwen2.5-7B & 72.80\%          & 72.80\%          & 72.80\%          & 70.06\%     & 54.55\%          & 72.12\%          & \textbf{73.42\%} 
\\
\midrule
Avg. & Qwen2.5-3B & 67.22\%          & 69.26\%          & 69.99\%          & 65.73\%     & 67.60\%          & 70.52\%          & \textbf{71.93\%}\\
& Qwen2.5-7B & 70.76\%          & 72.04\%          & 73.97\%          & 71.34\%     & 70.71\%          & 73.57\%          & \textbf{76.18\%}\\
\bottomrule
\end{tabular}
}
\caption{1-shot}

    \label{tab:subtable1}
\end{subtable}
\hspace{0.1cm}
\begin{subtable}{0.45\textwidth}
    \centering
    \resizebox{\textwidth}{!}{
\begin{tabular}{@{\extracolsep{\fill}} l c c c c c c c c}
\toprule
\multirow{2}{*}{\textbf{DataSet}} & \multirow{2}{*}{\textbf{LM}} & \multicolumn{7}{c}{\textbf{Method}} \\
 & &  \textbf{ICL} & \textbf{BC} & \textbf{LinC} & \textbf{CC+} & \textbf{BC+} & \textbf{LinC+} & \textbf{Ours} \\
\midrule
SST-2 & Qwen2.5-3B & 83.52\%          & 83.64\%          & 84.35\%          & 83.14\%          & 85.17\% & \textbf{85.83\%} & 81.22\%\\
& Qwen2.5-7B & 94.28\%          & 94.28\%          & 94.34\%          & \textbf{94.67\%} & 94.61\%          & 94.56\%          & 94.89\%\\
\midrule
MNLI & Qwen2.5-3B & 63.68\%          & 67.92\%          & 69.83\%          & 64.30\%          & \textbf{77.96\%} & 64.72\%          & 75.33\%\\
& Qwen2.5-7B & 64.41\%          & 70.79\%          & 73.30\%          & 75.42\%          & \textbf{82.58\%} & 65.66\%          & 79.82\%\\
\midrule
MRPC & Qwen2.5-3B & 66.09\%          & 63.13\%          & 67.54\%          & 71.88\%& 70.60\%          & 69.62\%          & \textbf{72.17\%}\\
& Qwen2.5-7B & 66.96\%          & 66.32\%          & 70.49\%          & 72.99\%          & \textbf{74.43\%} & 74.37\%          & 72.63\%\\
\midrule
QNLI & Qwen2.5-3B & 70.64\%          & 73.18\%          & 73.00\%          & 65.75\%          & \textbf{78.47\%} & 78.34\%          & 77.41\%\\
& Qwen2.5-7B & 78.64\%          & 78.86\%          & 78.74\%          & 71.15\%          & \textbf{80.57\%} & 80.34\%          & 80.01\%\\
\midrule
RTE & Qwen2.5-3B & 73.65\%          & 76.53\%          & 76.53\%          & 71.12\%          & 75.09\%          & 75.45\% & \textbf{78.70\%}\\
& Qwen2.5-7B & 80.87\%          & 81.23\%          & 80.87\%          & 75.09\%          & 77.25\%          & 79.06\%          & \textbf{81.59\%}\\
\midrule
WiC & Qwen2.5-3B & 58.00\%          & \textbf{58.35\%} & 57.57\%          & 53.92\%          & 58.14\%          & 58.29\% & 58.14\%          
\\
& Qwen2.5-7B & 61.00\%          & 62.35\%          & 62.36\%          & 58.14\%          & \textbf{63.07\%} & 63.00\% & 60.07\%          
\\
\midrule
YouTube & Qwen2.5-3B & 88.52\%          & 88.52\% & 88.27\%          & 85.97\%          & 76.02\%          & 88.26\%          & \textbf{88.78\%}\\
& Qwen2.5-7B & 88.77\% & 89.28\% & 89.03\% & 79.84\%          & 72.45\%          & \textbf{91.32}\% & 89.76\%\\
\midrule
AIGA & Qwen2.5-3B & 74.09\% & 73.90\%          & 73.92\%          & 76.55\%          & 72.15\%          & 72.78\%          & \textbf{74.39\%} 
\\
& Qwen2.5-7B & 74.93\%          & 75.10\%          & 74.98\%          & 77.23\%          & 77.93\%          & \textbf{79.65\%} & 75.83\%
\\
\midrule
Avg. & Qwen2.5-3B & 72.27\%          & 73.15\%          & 73.88\%          & 71.58\%          & 74.20\%          & 74.16\%          & \textbf{75.76\%}\\
& Qwen2.5-7B & 76.23\%          & 77.28\%          & 78.01\%          & 75.57\%          & 77.86\%          & 78.50\%          & \textbf{79.33\%}\\
\bottomrule
\end{tabular}}
    \caption{2-shot}
    \label{tab:subtable2}
\end{subtable}
\\[1ex]
\begin{subtable}{0.45\textwidth}
    \centering
    \resizebox{\textwidth}{!}{
  \begin{tabular}{@{\extracolsep{\fill}} l c c c c c c c c}
\toprule
\multirow{2}{*}{\textbf{DataSet}} & \multirow{2}{*}{\textbf{LM}} & \multicolumn{7}{c}{\textbf{Method}} \\
 & &  \textbf{ICL} & \textbf{BC} & \textbf{LinC} & \textbf{CC+} & \textbf{BC+} & \textbf{LinC+} & \textbf{Ours} \\
\midrule
SST-2 & Qwen2.5-3B & 89.68\%          & 89.79\%          & 89.90\%          & 90.72\%          & \textbf{91.27\%} & 91.21\% & 90.11\%\\
& Qwen2.5-7B & 95.17\%          & 95.06\%          & 95.06\%          & 95.22\% & 95.22\%          & \textbf{95.27\%} & 95.06\%\\
\midrule
MNLI & Qwen2.5-3B & 68.57\%          & 74.64\%          & 75.57\%          & 71.66\%          & \textbf{78.86\%} & 69.15\%          & 78.15\%\\
& Qwen2.5-7B & 69.29\%          & 77.34\%          & 79.08\%          & 80.74\%          & \textbf{84.05\%} & 69.32\%          & 82.54\%\\
\midrule
MRPC & Qwen2.5-3B & 69.80\%          & 67.54\%          & 69.22\%          & 70.84\% & 71.07\%          & 71.01\%          & \textbf{71.82\%}\\
& Qwen2.5-7B & 71.07\%          & 70.08\%          & 73.39\%          & 74.08\%          & \textbf{74.67\%} & 74.20\%          & 74.37\% 
\\
\midrule
QNLI & Qwen2.5-3B & 74.83\%          & 77.83\%          & 77.74\%          & 71.79\%          & \textbf{79.64\%} & 79.18\%          & 78.89\%\\
& Qwen2.5-7B & 79.95\%          & 80.08\%          & 80.17\%          & 75.32\%          & \textbf{81.18\%} & 81.22\%          & 80.62\%\\
\midrule
RTE & Qwen2.5-3B & 77.98\%          & 85.56\%          & 85.19\%          & 76.53\%          & 78.34\%          & 78.34\% & \textbf{85.92\%}\\
& Qwen2.5-7B & 84.11\%          & 84.11\%          & \textbf{84.84\%} & 78.70\%          & 83.03\%          & 84.83\%          & 84.11\%\\
\midrule
WiC & Qwen2.5-3B & 56.64\%          & 56.85\% & 57.21\%          & 55.36\%          & \textbf{58.93\%} & 58.71\% & 56.07\%\\
& Qwen2.5-7B & 62.42\%          & 62.42\%          & 62.42\%          & 60.28\%          & \textbf{63.85\%} & 63.57\% & 62.64\%\\
\midrule
YouTube & Qwen2.5-3B & 86.48\%          & 86.22\% & 86.48\%          & 82.40\%          & 80.10\%          & 89.28\%          & \textbf{90.05\%}\\
& Qwen2.5-7B & 89.28\% & 89.03\% & 89.03\% & 86.48\%          & 86.48\%          & 88.52\% & \textbf{90.82\%}\\
\midrule
AIGA & Qwen2.5-3B & \textbf{77.26\%} & 78.82\%          & 78.80\%          & 75.81\%          & 76.91\%          & 78.24\%          & \textbf{79.43\%}\\
& Qwen2.5-7B & 77.68\%          & 79.79\%          & 80.37\%          & 77.72\%          & 81.24\%          & \textbf{82.50\%} & 80.72\%\\
\midrule
Avg. & Qwen2.5-3B & 75.16\%          & 77.16\%          & 77.51\%          & 74.39\%          & 76.89\%          & 76.89\%          & \textbf{78.81\%}\\
& Qwen2.5-7B & 78.62\%          & 79.74\%          & 80.55\%          & 78.57\%          & 81.22\%          & 79.93\%          & \textbf{81.36\%}\\
\bottomrule
\end{tabular}}
    \caption{4-shot}
    \label{tab:subtable3}
\end{subtable}
\hspace{0.1cm}
\begin{subtable}{0.45\textwidth}
    \centering
    \resizebox{\textwidth}{!}{
\begin{tabular}{@{\extracolsep{\fill}} l c c c c c c c c}
\toprule
\multirow{2}{*}{\textbf{DataSet}} & \multirow{2}{*}{\textbf{LM}} & \multicolumn{7}{c}{\textbf{Method}} \\
 & &  \textbf{ICL} & \textbf{BC} & \textbf{LinC} & \textbf{CC+} & \textbf{BC+} & \textbf{LinC+} & \textbf{Ours} \\
\midrule
SST-2 & Qwen2.5-3B & 90.22\%          & 90.28\%          & 90.44\%          & 91.49\%          & \textbf{91.59\%} & 91.37\% & 90.44\%\\
& Qwen2.5-7B & 94.12\%          & 94.17\%          & 94.34\%          & 94.23\% & 94.34\%          & 94.34\% & \textbf{94.40\%} 
\\
\midrule
MNLI & Qwen2.5-3B & 69.25\%          & 75.87\%          & 76.55\%          & 72.65\%          & 78.63\%& 70.42\%          & \textbf{78.64\%}\\
& Qwen2.5-7B & 70.33\%          & 78.19\%          & 79.03\%          & 82.13\%          & \textbf{84.60\%} & 71.40\%          & 83.26\%\\
\midrule
MRPC & Qwen2.5-3B & 71.30\%          & 69.39\%          & 70.31\%          & 70.72\% & 71.30\%          & 71.07\%          & \textbf{72.17\%}\\
& Qwen2.5-7B & 72.00\%          & 70.26\%          & 72.99\%          & 73.45\%          & \textbf{74.55\%} & 73.57\%          & 73.85\%\\
\midrule
QNLI & Qwen2.5-3B & 76.20\%          & 77.81\%          & 77.81\%          & 71.05\%          & 79.17\% & \textbf{79.39\%} & 78.47\%\\
& Qwen2.5-7B & 79.90\%          & 79.71\%          & 79.79\%          & 73.09\%          & \textbf{81.15\%} & 81.11\%          & \textbf{80.23\%}\\
\midrule
RTE & Qwen2.5-3B & 80.14\%          & \textbf{85.20\%} & \textbf{85.20\%} & 74.73\%          & 79.42\%          & 79.78\% & \textbf{85.20\%} 
\\
& Qwen2.5-7B & 84.11\%          & 84.48\%          & \textbf{84.84\%} & 79.42\%          & 83.39\%          & 83.39\%          & \textbf{84.48\%}\\
\midrule
WiC & Qwen2.5-3B & 57.50\%          & 57.35\% & 57.28\%          & 56.42\%          & \textbf{58.64\%} & 58.14\% & 57.71\%\\
& Qwen2.5-7B & 62.64\%          & 62.07\%          & 62.50\%          & 62.43\%          & \textbf{64.00\%} & 63.28\% & 63.71\%\\
\midrule
YouTube & Qwen2.5-3B & 84.94\%          & 88.26\% & 88.26\%          & 87.86\%          & 79.59\%          & 87.76\%          & \textbf{91.83\%}\\
& Qwen2.5-7B & 87.76\% & 89.80\% & 89.54\% & 87.50\%          & 86.73\%          & 90.31\% & \textbf{91.07\%}\\
\midrule
AIGA & Qwen2.5-3B & 78.90\% & 79.88\%          & 80.01\%          & 74.86\%          & 77.81\%          & 79.29\%          & \textbf{82.86\%}\\
& Qwen2.5-7B & 79.50\%          & 81.86\%          & 82.01\%          & 77.05\%          & 80.75\%          & 82.74\% & \textbf{82.92\%}\\
\midrule
Avg. & Qwen2.5-3B & 76.06\%          & 78.01\%          & 78.23\%          & 74.97\%          & 77.02\%          & 77.15\%          & \textbf{79.66\%}\\
& Qwen2.5-7B & 78.80\%          & 80.07\%          & 80.63\%          & 78.66\%          & 81.19\%          & 80.02\%          & \textbf{81.74\%}\\
\bottomrule
\end{tabular}}
    \caption{5-shot}
    \label{tab:subtable4}
\end{subtable}
\caption{Accuracy (\%) comparison of different calibration methods on various datasets using BM25 selection strategy, increase ordering strategy, and Qwen2.5 models (3B and 7B) with 1- to 5-shot settings. The best performance for each dataset and model size is highlighted in bold.}

\label{tab:main}
\end{table*}






\end{document}